%% file: root.tex
\documentclass[letterpaper, 10 pt, conference]{ieeeconf}  %

\IEEEoverridecommandlockouts                              %

\usepackage{amsmath}
\usepackage{hyperref}
\usepackage{tikz}
\usetikzlibrary{shapes,arrows, calc, arrows.meta, intersections, positioning, patterns, decorations.pathreplacing, backgrounds} %
\usepackage{selectp} %
\usepackage{mathtools}
\usepackage{subcaption}
\usepackage{amsfonts}
\usepackage[noend]{algorithm2e}
\usepackage{calculator}
\usepackage{siunitx}
\usepackage[absolute,overlay]{textpos}
\usepackage{cite}
\usepackage{makecell, multirow}
\usepackage{comment}
\setcellgapes{1.1pt}
\usepackage{booktabs}
\newcommand{\tabitem}{~~\llap{\textbullet}~~}

\newcommand*\rot{\rotatebox{90}}
\definecolor{TABLE_GOOD}{RGB}{0,100,0}
\definecolor{TABLE_BAD}{RGB}{214,39,40}
\definecolor{PATH_OKAY}{RGB}{0,0,0}
\definecolor{PATH_COLLISION}{RGB}{230,0,0}
\definecolor{PLOT_RED}{RGB}{214,39,40}
\definecolor{PLOT_BLUE}{RGB}{0,0,150}
\definecolor{LINE_COLOR_RED}{RGB}{214,39,40}
\definecolor{LINE_COLOR_BLUE}{RGB}{0,0,150}
\definecolor{LINE_COLOR_GREEN}{RGB}{44,160,44}
\def\linkToCode{www.github.com/translearn/safeMotions}
\RestyleAlgo{ruled}
\DeclareMathOperator{\trajectory}{trajectory}

\title{\LARGE \bf
Learning Collision-free and Torque-limited Robot Trajectories based on Alternative Safe Behaviors
}

\author{Jonas C. Kiemel$^{1}$ and Torsten Kröger %
\thanks{\protect\hypertarget{link:author}{$^{1}$}Institute for Anthropomatics and Robotics – Intelligent Process Automation and Robotics (IAR-IPR), Karlsruhe Institute of Technology (KIT), jonas.kiemel@kit.edu \newline
	\indent $^{2}$\url{\linkToCode} 
	}%
}

\begin{document}

\maketitle
\thispagestyle{empty}
\pagestyle{empty}

\begin{textblock*}{14.9cm}(3.2cm,0.75cm) 
	{\footnotesize © 2022 IEEE.  Personal use of this material is permitted.  Permission from IEEE must be obtained for all other uses, in any current or future media, including reprinting/republishing this material for advertising or promotional purposes, creating new collective works, for resale or redistribution to servers or lists, or reuse of any copyrighted component of this work in other works.}
\end{textblock*}

\begin{abstract}

This paper presents an approach for learning online generation of collision-free and torque-limited robot trajectories. 
In order to generate future motions, a neural network is periodically invoked.  
Based on the current kinematic state of the robot and the network output, a trajectory for the current time interval can be calculated. 
The main idea of our paper is to execute the computed motion only if a collision-free and torque-limited way to continue the trajectory is known.
In practice, the motion computed for the current time interval is extended by a braking trajectory and simulated using a physics engine. 
If the simulated trajectory complies with all safety constraints, the computed motion is carried out.  %
Otherwise, the braking trajectory calculated in the previous time interval serves as an alternative safe behavior. 
Given a task-specific reward function, the neural network is trained using reinforcement learning.
The design of the action space used for reinforcement learning ensures that all computed trajectories comply with kinematic joint limits.  
For our evaluation, simulated humanoid robots and industrial robots are trained to reach as many randomly placed target points as possible. 
We show that our method reliably prevents collisions with static obstacles and collisions between the robot arms, while generating motions that respect both torque limits and kinematic joint limits.
Experiments with a real robot demonstrate %
that safe trajectories can be generated in \mbox{real-time}.

\end{abstract}

\section{INTRODUCTION}
Although robots are widely used for industrial automation purposes, many everyday tasks cannot yet be automated.
While it seems natural for humans to perform motions in constantly changing environments, the usage of robots is often limited to repetitive tasks in well-defined environments.
Generating fast and adaptive motions through reinforcement learning (RL) is a promising approach to overcome repetitive motion sequences and to automate more complex tasks in a human-like manner. %

However, in order to learn well-performing motions, robots must be able to explore their environment, 
At the beginning of a training process, this typically involves performing random motions.
The arms of a humanoid robot are likely to collide if random motions are performed. 
The same applies to industrial robots that share the same workspace in a production line.
In order to prevent damage to the robots and their environment, collisions have to be avoided
during the training process. 
Aside from collisions, a robot can also be damaged if one of its joints is overloaded. 
The kinematic capabilities of a robot are typically limited by the maximum \mbox{position $\theta$}, \mbox{velocity $\dot{\theta}$}, \mbox{acceleration $\ddot{\theta}$} and \mbox{jerk $\dddot{\theta}$} of each joint, whereas the dynamic potential is restricted by the maximum \mbox{torque $\tau$} of each actuator. %
In this work, we consider a motion as safe, if no collision occurs and if no kinematic joint limit or torque limit is violated.
\begin{figure}[t]
\captionsetup[subfigure]{margin=80pt}
    \vspace{0.2cm}
	
	    \begin{subfigure}[c]{0.23\textwidth}
	   \vspace{0.0cm} 
	   \includegraphics[trim=185 200 1035 240, clip, width=\textwidth]{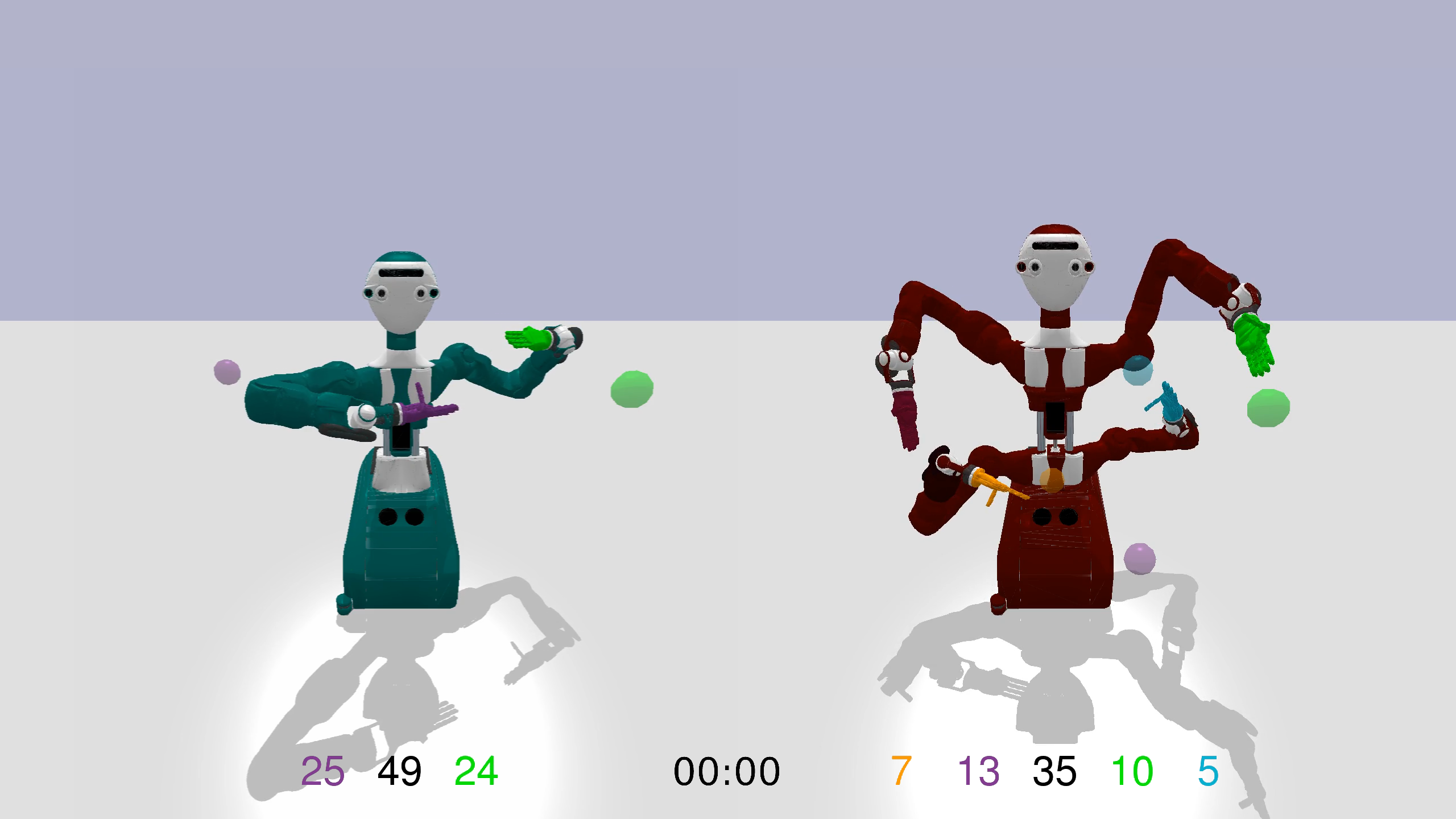}
	   
	   \vspace{-0.425cm}\hspace*{-2.05cm}\subcaptionbox{ARMAR-6}[8cm]

	\end{subfigure}
	\hspace{0.0065\textwidth}
	\begin{subfigure}[c]{0.23\textwidth}
	    \vspace{0.0cm}
	    \includegraphics[trim=1060 200 160 240, clip, width=\textwidth]{figures/TS_last_pic.png}
	    
		\vspace{-0.425cm}\hspace*{-2.55cm}\subcaptionbox{ARMAR-6x4}[9cm]

	\end{subfigure} 
	
	\vspace{0.18cm}
	
	    \begin{subfigure}[c]{0.23\textwidth}
	   \vspace{-0.0cm}
	   \includegraphics[trim=610 200 610 240, clip, width=\textwidth]{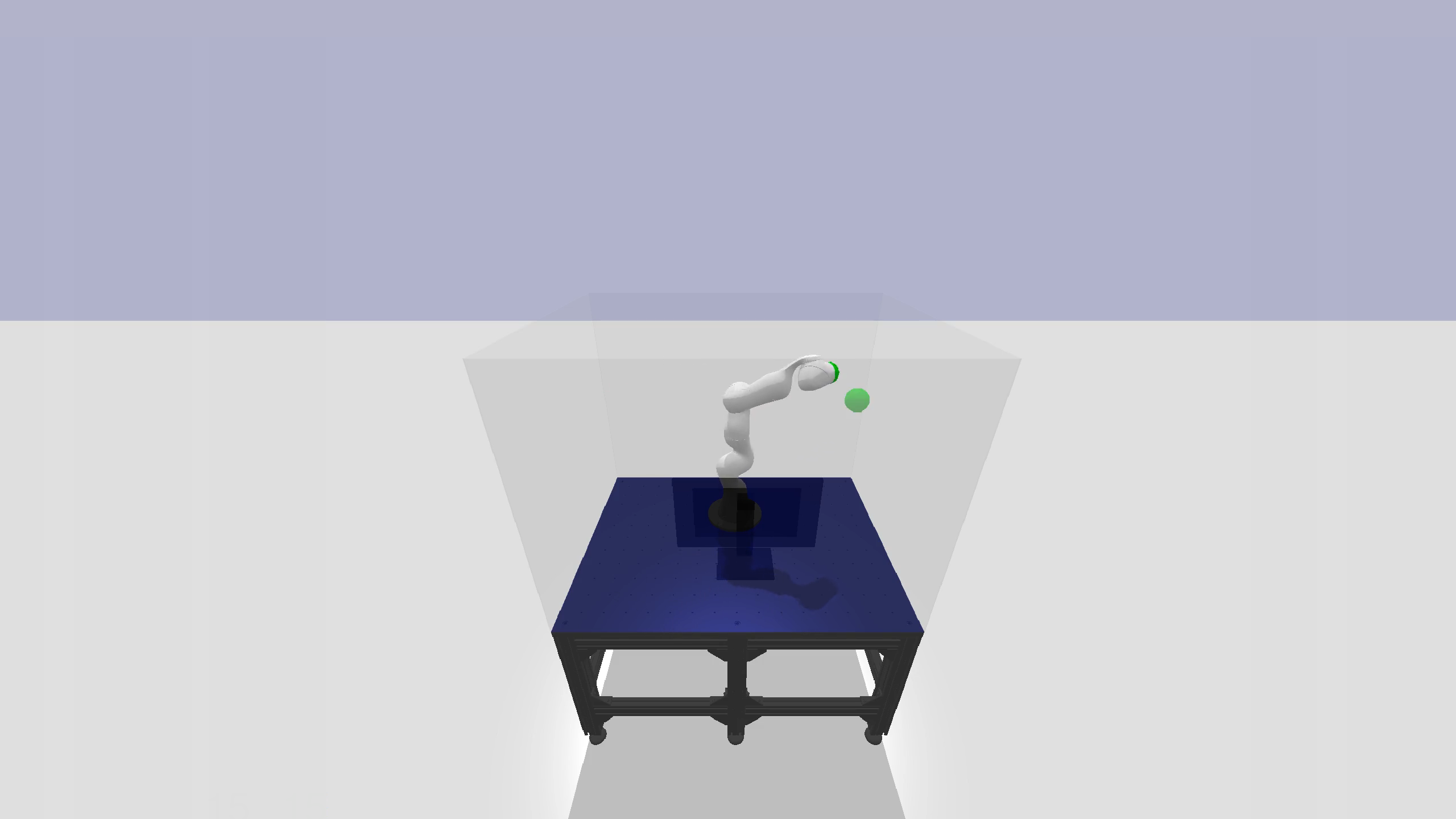}

	   \vspace{-0.38cm}\hspace*{-2.5cm}\subcaptionbox{One industrial robot}[9cm]

	\end{subfigure}
	\hspace{0.0065\textwidth}
	\begin{subfigure}[c]{0.23\textwidth}
	    \vspace{-0.0cm}
	    \includegraphics[trim=610 200 610 240, clip, width=\textwidth]{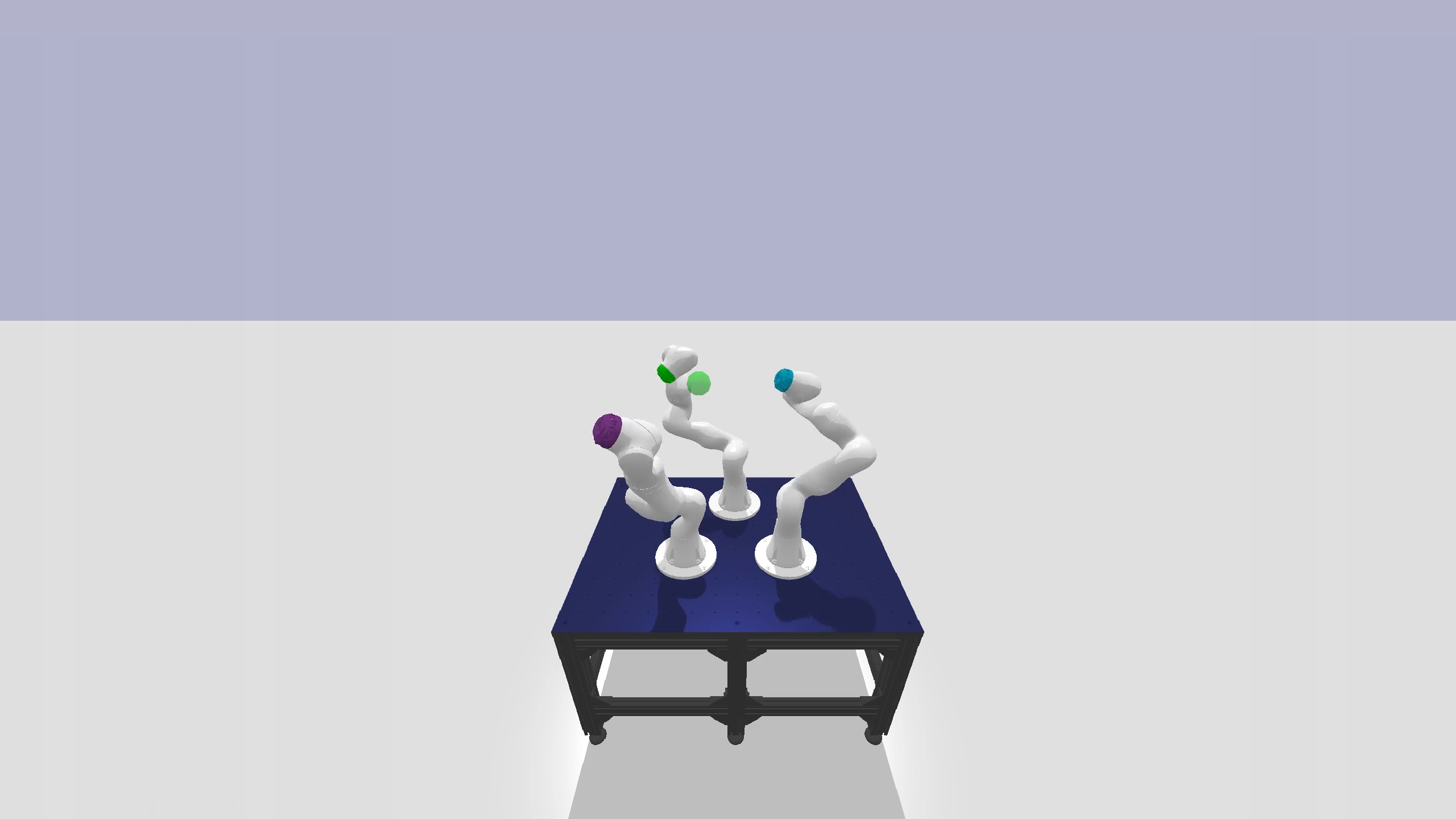}
	    
		\vspace{-0.38cm}\hspace*{-2.52cm}\subcaptionbox{Three industrial robots}[9cm]

	\end{subfigure}

	\caption{%
Our environments for learning safe motions.}%
	\label{fig:evaluation_environments_humanoid}
	\vspace{-0.32cm}
\end{figure}
For online trajectory generation, however, it is not sufficient to check whether all safety constraints are met within the current decision step as inevitable safety violations might occur at a later point in time. 
Consequently, a robotic system must check compliance with safety constraints over an infinite time-horizon or until a safe goal state is \mbox{found \cite{ShortPaperMotionSafety}}.  %
A major difficulty in this context is that future motions cannot be fully calculated and checked for safety violations in advance, as they depend on unknown future states of the environment.
To overcome this problem, two ideas \mbox{are integrated in our work}:
\begin{itemize}
    \item Compliance with kinematic joint limits is ensured over an infinite time-horizon by the design of the action space used for reinforcement learning.
    \item Collisions and torque limit violations are prevented by ensuring the existence of an alternative safe behavior at each decision step.
\end{itemize}
As a consequence, a multi-armed humanoid or a system of several industrial robots will execute safe motions only.
Thus, a neural network can be used to learn movements without having to worry about potential safety violations. 

Our main contributions can be summarized as follows: 
\begin{itemize}
\item We introduce an approach to learn safe motions for robotic manipulators with explicit safety guarantees during the entire training process considering kinematic joint limits, torque limits and collisions between multiple moving robots. 
\item We demonstrate the effectiveness and \mbox{scalability of} our method by learning fast yet safe motions in complex environments with up to four robot arms and more than 30 degrees of freedom \mbox{(see Fig. \ref{fig:evaluation_environments_humanoid}).} %
\item We show that our approach enables safe motions to be computed in real-time and demonstrate successful \mbox{sim-2-real} transfer with an industrial robot.
\end{itemize}

In addition, our source code and our trained neural networks have been made publicly available.\href{https://\linkToCode}{$^2$}

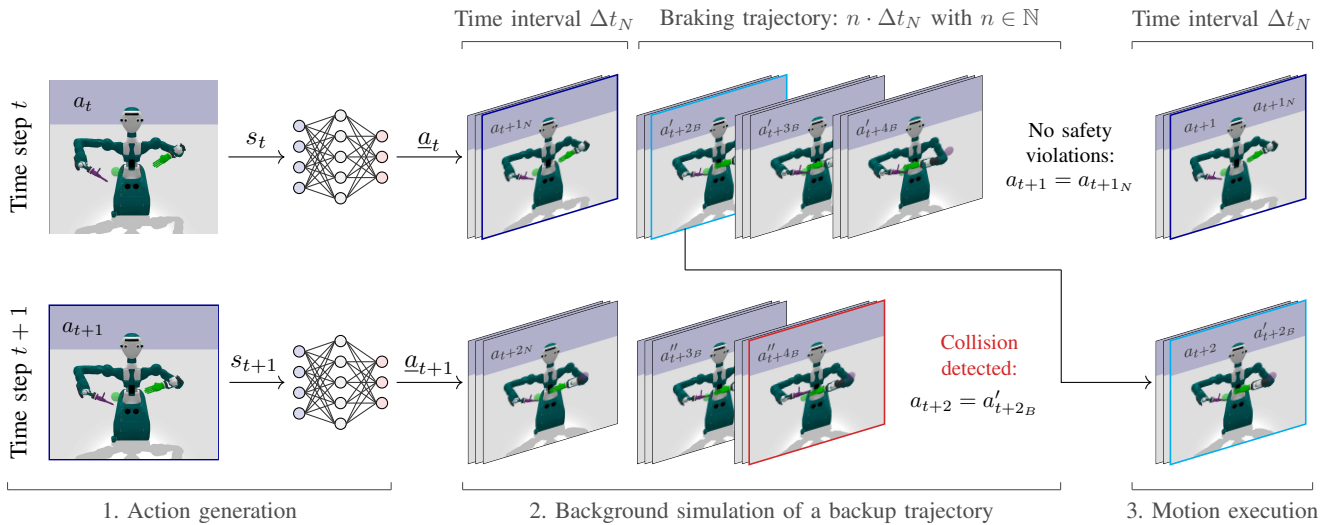
\begin{figure*}[t]
    \input{figures/asb_demo}
    \vspace{-0.1cm}
	\caption{%
	Our approach explained based on two exemplary time steps:  In the upper case, no safety violation is detected during a background simulation. Thus, the time step is executed as determined by the neural network. In the lower case, a collision is detected, which is why a safe braking acceleration from the previous background simulation is selected for execution.}
	\label{fig:basic_principle}
	\vspace{-0.4cm}
\end{figure*}

\section{Related work}

\subsection{Safe reinforcement learning}
The term safe reinforcement learning refers to methods that aim at respecting safety constraints during the learning phase and the deployment process. %
One important aspect of safe reinforcement learning is the problem of safe exploration. 
In order to find well-performing policies, an RL agent has to explore its environment. During this process, dangerous situations have to be avoided.
To account for safety constraints, the concept of Constrained Markov decision processes (CMDP) was introduced \cite{altman1999constrained}. 
CMDPs can be used to address two different types of constraints: Instantaneous constraints, which must be met at each time step, and cumulative constraints, which require the sum of a constraint penalty over time to be less than a specified threshold.
In~\cite{liu2021robot}, instantaneous constraints are considered by restricting the exploration of an RL agent on the tangent space of a constraint manifold.
Cumulative constraints are addressed by algorithms like Constrained  Policy  Optimization (CPO) \cite{achiam2017constrained},  Interior-point Policy Optimization (IPO) \cite{liu2020ipo} or SAMBA \cite{cowen2020samba}.
While constrained RL algorithms reduce the likelihood of constraint violations, they typically do not guarantee strict constraint satisfaction, especially at the beginning of the training process where agents behave randomly. %
Constraint satisfaction can be enforced by correcting unsafe actions, for instance with a linearized model trained on past trajectories~\cite{dalal2018safe}, using temporal logic specifications \cite{alshiekh2018safe} or by solving a constrained quadratic problem (QP) \cite{pham2018optlayer}. However, action correction mechanisms fail in states in which no safe action exists. 
Such a situation can be avoided by using control barrier functions (CBFs) \cite{ames2014control, wang2017safe, cheng2019end}, controlled invariant sets (CISs) \cite{anevlavis2021controlled, pannocchi2021trust} or by ensuring the existence of a safe backup trajectory at each decision step \cite{rubrecht2012motion, koller2018learning}.

\subsection{Learning safe motions for robotic manipulators}
When learning online trajectory generation with robotic manipulators, relevant safety constraints include collision avoidance and compliance with kinematic and dynamic joint limits.
Practitioners have developed various techniques to avoid unsafe behaviors when performing real-world experiments.  
For instance, penalties for undesired behaviors can be added to the reward function of an unconstrained Markov decision process (MDP) \cite{tan2018sim}.  %
While penalties reduce the likelihood of undesirable behaviors,
they have no effect at the beginning of a training process and do not provide strict safety guarantees thereafter. %
In some cases, task-specific heuristics can be used to avoid unsafe behaviors~\cite{gu2017deep}.
Designing the action space in such a way that all actions can be performed safely is another approach to address safety constraints. %
An action space representation that ensures compliance with kinematic joint constraints is presented in \cite{kiemel2020learning}. 
Conflicting constraints are avoided over an infinite time-horizon without restricting the workspace of the robot. 
When considering kinematic joint constraints only, all joints can be treated as decoupled. 
However, preventing collisions and torque limit violations is more challenging as the coupling between the joints has to be taken into account. %
Faverjon and Tournassoud’s method \cite{faverjon1987local} provides a way to avoid collisions by specifying explicit inequality constraints. 
In \cite{pham2018optlayer}, Faverjon and Tournassoud’s method is used to correct unsafe actions by solving a constrained quadratic problem (QP). %
However, the approach does not ensure that a safe action exists at every decision step.
In case of conflicting constraints, the QP has no solution and compliance with the safety constraints is no longer guaranteed. %
To avoid conflicting constraints when tracking Cartesian velocities with a hydraulic manipulator, the concept of alternative safe behaviors was proposed in \cite{rubrecht2012motion}.
Based on the idea of alternative safe behaviors, we introduce an approach that enables safe reinforcement learning of fast and smooth motions performed by robotic manipulators. %
Compared to~\cite{rubrecht2012motion}, we additionally consider torque constraints, jerk limits and collisions between multiple robots. 
Our approach provides explicit safety guarantees right from the beginning of the training process and continues to do so even if the environment is modified at a later time.
Furthermore, and unlike previous work, we show that our method can be applied to complex multi-arm systems with more than \mbox{30 degrees of freedom without causing safety violations.}

\label{sec:citations}
\section{Problem statement}
Our work addresses the problem of learning collision-free robot motions that comply with kinematic and dynamic joint limits via model-free RL. 
Specifically, the following kinematic constraints are defined for each revolute and prismatic robot joint:
\begin{alignat}{3}
p_{min} &{}\le{}& \theta &{}\le{}& p_{max}  \label{eq:constraint_p} \\ 
v_{min} &{}\le{}& \dot{\theta} &{}\le{}& v_{max}  \label{eq:constraint_v}\\
a_{min} &{}\le{}& \ddot{\theta}&{}\le{}& a_{max}  \label{eq:constraint_a}  \\
j_{min} &{}\le{}& \dddot{\theta} &{}\le{}& j_{max},  \label{eq:constraint_j}
\end{alignat}
with $\theta$ being the joint position and $p$, $v$, $a$ and $j$ standing for position, velocity, acceleration and jerk, respectively. 
In addition, the torque limits of each joint have to be respected:
\begin{align}
    \tau_{min} {}\le{} \tau {}\le{}  \tau_{max}
\end{align}
Compliance with the safety constraints must be ensured over an infinite time horizon. It is well known, however, that
collision avoidance cannot be guaranteed in the presence of arbitrarily moving obstacles \cite{ShortPaperMotionSafety}. 
Therefore, our work focuses on self-collisions, collisions between moving, but centrally controlled robots and collisions with static obstacles.
Areas where moving obstacles are present can be avoided by restricting the workspace through virtual walls.
For our experiments, we assume that the position and the shape of each obstacle in the environment are known.
\section{Approach}
\subsection{Basic principle}
We explain the basic principle of our approach based on two subsequent decision steps shown in Fig. \ref{fig:basic_principle}. 
Given the current state of the environment ${s}$ as input, a neural network generates an action $\underline{a}$ at each decision step.
The action is used to compute a motion with a duration of $\Delta t{_N}$, the time between discrete decision steps. 
This motion might or might not be safe. To prevent safety violations, we extend the computed motion with a braking trajectory that leads to a full stop of the robot system. 
The resulting backup trajectory is checked for collisions and torque limit violations using a physics simulator. 
In time step $t$ shown in the upper part of Fig. \ref{fig:basic_principle}, no safety violations are detected and the motion that resulted from the network action is executed. 
Contrary to that, a collision of the robot hands is detected in time step $t+1$. 
For that reason, the first part of the braking trajectory computed in time step $t$ serves as an alternative behavior.
Note that the execution of the alternative behavior in time step $t+1$ is guaranteed to be safe as it was checked for safety violations in time step $t$.
Since a safe resting state is never left without knowing a safe way to stop the robot system again, the existence of an alternative safe behavior is guaranteed for each decision step. As part of the learning process, the  neural network can learn to avoid actions that lead to the selection of the alternative behavior. In the subsections that follow, we explain how the problem is formalized as a Markov decision process (\ref{sec:MDP}) and how actions are mapped to safely executable motions (\ref{sec:action_mapping}). In addition, we describe how the braking trajectories are calculated (\ref{sec:braking}), how the backup trajectory is checked for safety violations (\ref{sec:background_simulation}) and how the required computations can be carried out \mbox{in real-time (\ref{sec:real_time}).}
\subsection{Formalization as a Markov decision process}
\label{sec:MDP}
To learn motions with model-free RL, we define an unconstrained Markov decision process $(\mathcal{S}, \mathcal{A}, P_{\underline{a}}, R_{\underline{a}})$, where $\mathcal{S}$ is the state space,  $\mathcal{A}$ is the action space,  $P_{\underline{a}}$ are unknown transition probabilities and $R_{\underline{a}}$ is the immediate reward resulting from action $\underline{a}$. Note that actions are underlined to distinguish them from accelerations $a$.
Actions are generated by a neural network that is trained to maximize the sum of immediate rewards.
A state ${s_t} \in \mathcal{S}$ consists of the current kinematic setpoints of each robot joint (position~$p_t$, velocity~$v_t$, acceleration~$a_t$) and a task-specific part.
Given a state ${s_t}$ as input, a neural network outputs a single scalar $ \in [{-1}, 1]$  per joint as action $\underline{a}_t \in \mathcal{A}$.
The immediate reward results from a task-specific reward function, which specifies the optimization goal of the learning problem. The time between decision steps is constant and \mbox{referred to as $\Delta t{_N}$.}
\newcommand\mycommfont[1]{\scriptsize\ttfamily\textcolor{black!80}{#1}}
\SetCommentSty{mycommfont}
\SetKwComment{Comment}{/* }{ */}
\SetKwFor{At}{at each time step}{do}{end}
\begin{algorithm}[t]
\caption{Computing a safe value for $a_{t+1}$}\label{alg:one}
\At{t}{
compute $a_{t+1_N}$ based on the action $\underline{a}_t$\;
compute braking accelerations starting from $a_{t+1_N}$:  ($a_{t+2_B}$, $a_{t+3_B}$, \ldots)\; 
$\emph{\textrm{backup trajectory}} \gets \trajectory(a_{t}, a_{t+1_N}, a_{t+2_B}, a_{t+3_B}, \ldots)$\;
\eIf{\textrm{backup trajectory} \textbf{\emph{is}} safe}{
      $a_{t+1} \gets a_{t+1_N}$\;
    }{
    $a_{t+1} \gets a_{t+1_B}$\hspace{-0.26em}\ \Comment*[]{safe braking acceleration computed in a previous time step} 
    }
}
\label{algo:safe_acc}
\vspace{-0.0em}
\end{algorithm}
\vspace{0em}
\subsection{Mapping actions to motions}
\label{sec:action_mapping}
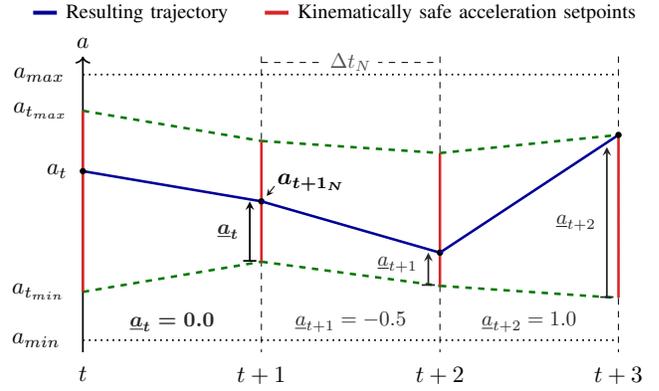
\begin{figure}[t]
	 \begin{tikzpicture}[auto, node distance=5cm,>=latex', scale=0.7,  every node/.style={scale=0.85}]
        \node[text width=4cm] at (0, 0) (origin){};
    	\draw [draw=LINE_COLOR_BLUE, line width=1.5pt] ($(origin.center)+(-2.0cm, -0.0cm)$) -- + (0.45cm, 0cm) node[pos=1, right, yshift=-0.01cm, align=left]{\small{Resulting trajectory}};
	    \draw [solid, draw=LINE_COLOR_RED, line width=1.5pt] ($(origin.center)+(2.4cm, 0.0cm)$) -- + (0.45cm, 0cm) node[pos=1, right, yshift=-0.01cm, align=left]{\small{Kinematically safe acceleration setpoints}};
    \end{tikzpicture} 
    
	\hspace{-5.5cm}
	\resizebox{0.8\textwidth}{!}{
    \input{figures/action_space}
    } %
	\caption{%
	The mapping from an action $\underline{a}_t$ to a kinematically feasible acceleration $a_{{t+1}_{N}}$ illustrated for a single joint.}
	\vspace{-0.7cm}
	\label{fig:action_space}
\end{figure}

At each decision step $t$, a safe motion with a duration of $\Delta t_N$ has to be generated. %
As described in Algorithm \ref{algo:safe_acc}, this is done by selecting a safe value for $a_{t+1}$, the desired joint acceleration at time step $t+1$.
Controlling a robot in joint space offers the advantage that neither singularities nor kinematic redundancies require special consideration.
To generate a continuous trajectory between the discrete decision steps, a linear interpolation between the joint accelerations $a_{t}$ and $a_{t+1}$ is performed. The corresponding joint velocities and joint positions can then be calculated by integration.
\newpage
\vspace*{-0.9cm}
As a first step towards finding a suitable value for $a_{t+1}$, a kinematically safe acceleration $a_{t+1_N}$ is computed based on the selected action $\underline{a}_t$. In the case of kinematic joint constraints, each joint can be considered independently. The action $\underline{a}_t$ consists of a scalar $\in [{-1}, 1]$  per joint.
We use the method described in \cite{kiemel2020learning} to compute for each joint the minimum and maximum acceleration $a_{t+1_{min}}$ and $a_{t+1_{max}}$ complying with the kinematic constraints (\ref{eq:constraint_p}) - (\ref{eq:constraint_j}). 
All accelerations between $a_{t+1_{min}}$ and $a_{t+1_{max}}$ lead to kinematically feasible trajectories.
As visualized in Fig. \ref{fig:action_space},  
$a_{{t+1}_{N}}$ is computed by linearly mapping the action $\underline{a}_t$ onto the range specified by $a_{t+1_{min}}$ and $a_{t+1_{max}}$:
\begin{align}
a_{t+1_N} = a_{t+1_{min}} + \frac{1 + \underline{a}_t}{2} \cdot \left(a_{t+1_{max}} - a_{t+1_{min}}\right)
\end{align}
The \textbf{default behavior} is to set $a_{t+1}$ to $a_{t+1_N}$. %
However, if a collision or a torque limit violation is detected in a subsequent background simulation, an \textbf{alternative behavior} is applied. 
In this case, $a_{t+1}$ is set to a safe braking acceleration computed in one of the previous time steps. Details on the computation of braking accelerations are given in the following subsection.
\subsection{Computation of braking trajectories} %
\label{sec:braking}
As shown in Fig. \ref{fig:basic_principle}, the background simulation to detect safety violations is performed for the next time interval specified by $a_{t+1_N}$ and for a subsequent braking trajectory.
Once the robot is stopped, no further safety violations can occur. %
Consequently, the braking trajectory leads the robot to a safe goal state.
The duration of the braking trajectory depends on the acceleration $a_{t+1_N}$ and the corresponding velocity $v_{t+1_N}$ but it is always a multiple of $\Delta t_N$.
The shorter the duration of the braking trajectory, the less computational effort is required for the background simulation. 
Therefore, we calculate a braking trajectory that is time-optimal with respect to predefined kinematic joint limits:
\begin{alignat}{3}
a_{min, \,brake} &{}\le{}& \ddot{\theta}&{}\le{}& a_{max, \,brake} \label{eq:constraint_a_brake}\\
j_{min, \,brake} &{}\le{}& \dddot{\theta} &{}\le{}& j_{max, \,brake}\label{eq:constraint_j_brake}
\end{alignat}
Two peculiarities have to be considered when calculating the desired braking accelerations \mbox{$a_{t+2_B}$, $a_{t+3_B}$,  \ldots}.
Firstly, the braking trajectory is not allowed to violate the kinematic limits specified in (\ref{eq:constraint_p}) - (\ref{eq:constraint_j}). For that reason, we calculate the 
range of kinematically safe accelerations for each time step and clip the corresponding braking acceleration if it falls outside this range.
Secondly, the desired kinematic target state ($v=0$ and $a=0$) must be reached at a discrete time step. Otherwise, the interpolation between the decision steps leads to an oscillation around the target state. We refer to~\cite{kiemel2020trueadapt} for a detailed analysis and solution of this problem. %

\subsection{Background simulation to detect safety violations}
\label{sec:background_simulation}
Potential collisions and torque limit violations of the backup trajectory are detected by a background simulation performed with the physics engine PyBullet \cite{coumans2016pybullet}.

In order to detect \textbf{collisions}, we observe the closest distance between specified object pairs.
Obstacle-link pairs are defined to detect collisions between robot links and obstacles. Link-link pairs are used to detect self-collisions and collisions between different robots. %
The distance between the observed pairs is checked at discrete time points with an adjustable frequency $f_C$ using the \texttt{getClosestPoints()} function provided by PyBullet.
The backup trajectory is considered as unsafe, if the minimum distance between an observed object pair is smaller than a user-specified safety distance $d_S$. Note that the computational effort depends on the duration of the backup trajectory and the selected frequency $f_C$.
The computational effort can be reduced by selecting a lower frequency $f_C$.
The resulting inaccuracy can then be compensated by choosing a larger safety distance $d_S$.

To  detect  \textbf{torque  limit  violations},  the  execution  of the backup  trajectory  is  simulated  with a frequency of \mbox{$f_S=$ \SI{240}{\hertz}}, the default simulation frequency of PyBullet. %
At each simulation  step,  the  torque applied to the joints is read out and checked for compliance with the specified limits.
From a theoretical perspective, the torque $\tau$ required to follow a desired joint \mbox{acceleration $\ddot{\theta}$} is described by the dynamical model of rigid multi-bodies: 
\begin{align}
\tau = M(\theta) \, \ddot{\theta} + C(\theta, \dot{\theta}) \, \dot{\theta} + G(\theta) + \tau_{ext},
\end{align}
where $M$ is the mass matrix, $C(\theta, \dot{\theta}) \, \dot{\theta}$ are centrifugal and Coriolis forces, $G(\theta)$ represents gravity forces and $\tau_{ext}$ summarizes external contact forces. 
Consequently, torque limit violations can arise from high joint accelerations $\ddot{\theta}$ but also from collisions, which typically lead to high external forces $\tau_{ext}$. %
The \mbox{mass matrix $M$} depends on the current joint position $\theta$ but also on the inertial properties of the robot links. Thus, the simulation accuracy can be improved by determining accurate inertial properties based on system identification. %
In practice, the simulation error can also be compensated to a certain extent by subtracting a safety margin from the torque limits specified by the robot manufacturer.
TABLE \ref{table:collision_vs_torque} summarizes our method to detect collisions and torque limit violations. 
We note that our approach can be easily extended to other safety constraints such as limits on Cartesian velocities or contact forces. 
The basic principle is the same as for the torque limit detection, except that other properties have to be read out from the physics engine.

\subsection{Real-time execution}
\label{sec:real_time}
\begin{figure}[h]
    \vspace{-0.2cm}
    \begin{tikzpicture}[scale=1.0]
            \def\scaleFactor{0.85}
             \def\ymax{0.75}
        	\def\ymin{0}
        	\def\azero{0.6}
        	\def\xdelta{7.4}
        	\def\xmax{1*\xdelta + 0.7}
        
        	\def\actionPredictionPos{0.27}
        	\def\actionPredictionTextPos{0.39}
        	\def\safetyChecksPos{0.51}
        	\def\safetyChecksTextPos{0.74}
        	\def\ticklength{3pt}

        	\definecolor{POS_LIM_A}{RGB}{170,0,0} %
        	\definecolor{POS_LIM_B}{RGB}{0,0,150}
        	\definecolor{POS_LIM_C}{RGB}{0,90,0}
        	\definecolor{POS_LIM_D}{RGB}{197,43,11}
        	\definecolor{POS_LIM_E}{RGB}{35,0,74}
    
    		\draw [->,thick, name path=pathX] (-0.25, 0) -- (\xmax,0)  node[pos=\actionPredictionPos] (actionPrediction) {} node[pos=\actionPredictionTextPos] (actionPredictionText) {} node[pos=\safetyChecksPos] (safetyChecks) {} node[pos=\safetyChecksTextPos] (safetyChecksText) {}  node (xaxis) [right] {};
    		
        	\draw [-{Latex[scale=1.0]},thick, color=black] (0*\xdelta,\ymin) node[below=0.08, black, xshift=0.0cm, name=timestep_t, outer sep=2pt] {$t$} -- (0*\xdelta,\ymax); %
        	\draw [-{Latex[scale=1.0]},thick, color=black] (1*\xdelta,\ymin) node[below=0.1, black, xshift=0.0cm, name=timestep_t_plus_one] {$t\!+\!1$} -- (1*\xdelta,\ymax); %
        	
        	\draw[thick] ($(actionPrediction.center)+(0, \ticklength)$) -- ($(actionPrediction.center)-(0, \ticklength)$);
        	\draw[thick] ($(safetyChecks.center)+(0, \ticklength)$) -- ($(safetyChecks.center)-(0, \ticklength)$);

	        \node[color=POS_LIM_B, align=center,  scale=\scaleFactor] at ($(actionPredictionText.center)+(0,0.40cm)$) {Action \\generation};
        	\node[color=POS_LIM_A, align=center,  scale=\scaleFactor] at ($(safetyChecksText.center)+(-0.22cm, 0.40cm)$) {Background simulation \\ of the backup trajectory};
        	
        	\draw[dashed, draw=black!50] (timestep_t.345)  -- + (0.9*\xdelta, 0) node[pos=0.57, fill=white, scale=\scaleFactor] {\textcolor{black!50}{$\Delta t_N$}};
        	
        	\def\braceXOffset{0.00}
        	\def\braceYOffset{0.1}
        	
        	\draw[black,decorate, decoration={brace,amplitude=4pt}] ($(actionPrediction.center) + (0,\ymax + \braceYOffset)$) -- ($(\xdelta, 0) + (0,\ymax + \braceYOffset)$)  node[midway, above,yshift=4pt, xshift=0.0cm, scale=0.9]{Precalculation for time step $t\!+\!1$};

    	\end{tikzpicture}
     \vspace{-0.5cm}
    \caption{The background simulation is performed in advance.}
    \label{fig:precalculation}
    \vspace{-0.5cm}
\end{figure}
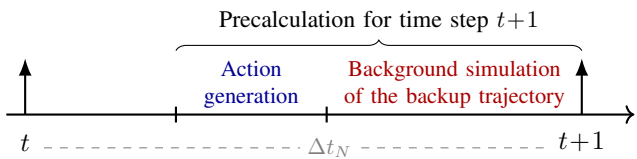

When generating motions for real robots, real-time requirements must be considered.  
As shown in Fig. \ref{fig:precalculation}, we perform the action generation and the background simulation ahead of the actual decision step $t+\!1$. 
For real-time execution, the computation time per decision step should never exceed~$\Delta t_N$. Under this condition, a safe motion is already known when the actual decision step $t+\!1$ is reached and the trajectory can be continued without interruption.

\begin{table}[t]
     \caption{Collision detection vs. torque limit detection}
     \vspace{-0.15cm}
    \makegapedcells
\begin{tabular*}{0.49\textwidth}{p{24mm}p{25.5mm}p{25.5mm}} 
    \toprule
\hspace{0.02cm}  & \multicolumn{1}{c}{Collision detection}  & \multicolumn{1}{c}{Torque limit detection} \\
    \hline
 Test frequency & \hfil $f_C \le f_S $ & \hfil $f_S = \SI{240}{\hertz}$ \\
\vspace{0.3cm}Criteria for unsafe \newline backup trajectories & \hspace{0.165cm} minimum distance \newline  \hspace*{0.18cm} \tabitem to obstacles \newline  \hspace*{0.18cm} \tabitem between links \newline \hspace*{0.35cm} smaller than a \newline \hspace*{0.24cm}  safety distance $d_S$ & \hfil torque limit violations \newline \null\hfil for instance due to  \newline  \hspace*{-0.0cm}\tabitem high accelerations 
\newline  \hspace*{-0.0cm}\tabitem gravity \newline  \hspace*{-0.0cm}\tabitem collisions \\
    \bottomrule
    \end{tabular*}
    \vspace{0ex}
\label{table:collision_vs_torque}
\vspace{-0.65cm}
\end{table}

\section{Evaluation}
\subsection{Environments used for evaluation}
We evaluate our approach with various humanoid robots and industrial robots %
shown in \mbox{Fig. \ref{fig:evaluation_environments_humanoid}}. 
Renderings are also provided in the accompanying video\footnote[3]{\url{https://youtu.be/U2OWsQrt-40}}. %
TABLE \ref{table:environments_metrics} summarizes the most important properties of each environment. 
The environment with a single `KUKA iiwa 7' robot (Fig. \ref{fig:evaluation_environments_humanoid}c) corresponds to our real-word setup shown in Fig. \ref{fig:real_robot_obstacle}. In this environment, the workspace of the robot is restricted by four virtual walls, a table and a monitor placed in front of the robot.  
Environments  with  two  and  three  industrial  robots  are  provided  to demonstrate  that  our  method  can  be  applied to multiple \mbox{robots sharing the same workspace.}
Applying our method to other robots is straightforward as the collision shapes and joint limits can be read out from robot descriptions that comply with the URDF standard. %
We also use our method to control both arms of the humanoid robot ARMAR-6 \cite{asfour2018armar} while avoiding self-collisions.
In order to demonstrate the scalability of our approach, we additionally introduce  `ARMAR-6x4', a four-armed modification of ARMAR-6 with 33 degrees of freedom. 
\begin{table}[t]
     \caption{Characteristics of our evaluation environments.}
     \vspace{-0.15cm}
    \makegapedcells
\begin{tabular*}{0.49\textwidth}{@{}p{20.4mm}p{3.5mm}p{8.0mm}p{21.3mm}p{16.8mm}} 
    \toprule
\hspace{0.02cm} Environment & DOF & Obstacles &  Obstacle-link pairs & Link-link pairs \\
    \hline
\hspace{0.02cm}    Industrial & & & & \\
\hspace{0.0002cm} \tabitem One robot  & \hfil \hspace{0.000cm} 7 & \hspace{0.000cm} \hfil \hspace{0.000cm} 6 & \hfil 36 & \hfil 0 \\
\hspace{0.0002cm} \tabitem Two robots & \hfil \hspace{0.000cm} 14 & \hspace{0.000cm} \hfil  \hspace{0.000cm} 1 & \hfil  12 & \hfil 72 \\
\hspace{0.0002cm} \tabitem Three robots & \hfil \hspace{0.000cm} 21 & \hspace{0.000cm} \hfil \hspace{0.000cm} 1 & \hfil 18 & \hfil  216 \\
\hline
\hspace{0.02cm}    Humanoids & & & & \\
\hspace{0.0002cm} \tabitem ARMAR-6  & \hfil \hspace{0.000cm} 17 &  \hspace{0.000cm} \hfil \hspace{0.000cm} 1 & \hfil 2 & \hfil 151 \\
\hspace{0.0002cm} \tabitem ARMAR-6x4 & \hfil \hspace{0.000cm} 33 & \hspace{0.000cm} \hfil  \hspace{0.000cm} 1 & \hfil  4 & \hfil 610 \\
    \bottomrule
    \end{tabular*}
    \vspace{0ex}
\label{table:environments_metrics}
\vspace{-0.55cm}
\end{table}
\begin{table*}[t]
    \caption{Evaluation of our collision prevention method based on 900 simulated episodes.}
    \vspace{-0.15cm}
    \makegapedcells
\begin{tabular*}{\textwidth}{@{}p{33.8mm}p{15mm}p{10mm}p{14mm}p{8mm}p{8mm}p{8mm}p{8mm}p{8mm}p{9.0mm}p{11.0mm}} 
    \toprule

\multirow[t]{3}{*}{\hspace{0.02cm} Configuration}
    & \multicolumn{3}{c}{Random agent} &  \multicolumn{7}{c}{Trained agent} \\
     & Target points & \multicolumn{1}{c}{Closest} & \multicolumn{1}{l|}{Adaptation}  & \multicolumn{5}{c}{Target points} & \multicolumn{1}{c}{Closest}  & \multicolumn{1}{c}{Adaptation} \\
     & \hfil All arms \hfil & \multicolumn{1}{c}{distance} & \multicolumn{1}{c|}{rate}  & \multicolumn{1}{c}{All arms} & Arm 1 & Arm 2 & Arm 3 & Arm 4 & \multicolumn{1}{c}{distance} & \multicolumn{1}{c}{rate}\\
    \hline
\hspace{0.02cm} One industrial robot & & & & \\
\tabitem single target point
    & \hfil $0.06$ \hfil & \hfil \SI{0.93}{\cm} \hfil   &  \hfil \SI{12.9}{\percent} \hfil  &  \hfil $8.90$    &  \hfil $8.90$ & \hfil -- & \hfil -- & \hfil -- &  \SI{0.88}{\cm} & \hfil \SI{1.6}{\percent} \\
      
     \hline %
\hspace{0.02cm} Two industrial robots  & & & & \\  
\tabitem simultaneous target points
    & \hfil $0.09$ & \hfil \SI{0.91}{\cm} &  \hfil \SI{8.1}{\percent}   &  \hfil $10.46$    & \hfil $2.08$ & \hfil $8.38$ & \hfil -- & \hfil -- &  \SI{0.62}{\cm} & \hfil \SI{3.1}{\percent} \\
\tabitem alternating target points
    & \hfil $0.04$ & \hfil\SI{0.93}{\cm} \hfil    &   \hfil \SI{8.1}{\percent}   &   \hfil $8.43$    & \hfil $4.21$ & \hfil $4.22$ & \hfil -- & \hfil -- &  \SI{0.65}{\cm} & \hfil \SI{2.3}{\percent} \\
     \hline%
\hspace{0.02cm} Three industrial robots  & & & & \\
\tabitem simultaneous target points
    & \hfil $0.10$ & \hfil\SI{0.82}{\cm} \hfil   &  \hfil \SI{18.4}{\percent}   &  \hfil $5.67$    & \hfil $4.46$ & \hfil $0.20$  & \hfil $1.01$ & \hfil -- &  \SI{0.72}{\cm} & \hfil \SI{17.2}{\percent} \\
\tabitem alternating target points
    & \hfil $0.04$ & \hfil \SI{0.91}{\cm} \hfil   &  \hfil \SI{18.6}{\percent}  &  \hfil $5.07$   & \hfil $1.68$ & \hfil $1.75$ & \hfil $1.64$ \hfil & \hfil -- &  \SI{0.71}{\cm} & \hfil \SI{5.1}{\percent} \\
     \bottomrule
\hspace{0.02cm} ARMAR-6 & & & & \\
\tabitem single target point
    & \hfil $0.06$ & \hfil  \SI{0.60}{\cm}    &  \hfil \SI{5.6}{\percent}     &  \hfil $13.06$    &  \hfil $6.33$ & \hfil $6.73$ & \hfil -- & \hfil -- &  \SI{0.95}{\cm} & \hfil \SI{1.7}{\percent} \\
\tabitem simultaneous target points
    & \hfil $0.05$ & \hfil \SI{0.79}{\cm} \hfil  & \hfil \SI{5.4}{\percent}   & \hfil $17.60$   & \hfil $7.86$ & \hfil $9.74$ & \hfil -- & \hfil -- &  \SI{0.21}{\cm} & \hfil \SI{7.0}{\percent} \\
\tabitem alternating target points
    & \hfil $0.02$ &  \hfil \SI{0.46}{\cm} \hfil&  \hfil \SI{5.5}{\percent} &  \hfil $12.40$    & \hfil $6.22$ & \hfil $6.18$ & \hfil -- & \hfil -- &  \SI{0.62}{\cm} & \hfil \SI{1.7}{\percent} \\  

    \hline %
\hspace{0.02cm}  ARMAR-6x4  & & & & \\  
\tabitem single target point
    & \hfil $0.06$ & \hfil \SI{0.51}{\cm} \hfil     &  \hfil \SI{22.8}{\percent}  &  \hfil $7.95$    &  \hfil 0.07 & \hfil $0.01$ & \hfil $4.09$ & \hfil $3.78$ &  \SI{0.32}{\cm} & \hfil \SI{7.3}{\percent} \\
\tabitem simultaneous target points
    & \hfil $0.10$ & \hfil \SI{0.45}{\cm}    &  \hfil \SI{22.2}{\percent}    & \hfil $7.37$    & \hfil $2.24$  & \hfil $2.08$ & \hfil $1.26$ & \hfil $1.79$ &  \SI{0.19}{\cm} & \hfil \SI{27.2}{\percent} \\
\tabitem alternating target points
    & \hfil $0.03$ & \hfil \SI{0.73}{\cm} \hfil   &   \hfil \SI{22.8}{\percent}  & \hfil $5.53$    & \hfil $1.37$ & \hfil $1.41$ & \hfil $1.36$ & \hfil $1.39$ &  \SI{0.38}{\cm} & \hfil \SI{7.4}{\percent} \\  
    \bottomrule
    \end{tabular*}
\label{table:collision_prevention}
\vspace{-0.33cm}
\end{table*}

\subsection{Details on the learning task and the training process}
Each of the robot arms is trained to reach as many randomly placed target points as possible within a fixed amount of time. %
We evaluate three different variants to assign target points to the end effectors of the robot arms: In case of simultaneous target points, each robot arm has a dedicated target point. 
With alternating target points, one target point is assigned to the robot arms in alternating order. `Single target point' means that a single target point is assigned to all robot arms. 
The duration of each episode is set to eight seconds.
As soon as a target point is reached, a new one is placed. 
The task-specific part of the state ${s} \in \mathcal{S}$ contains the Cartesian position of each target point and the relative position between the target points and their corresponding robot arms. 
Note that the position of obstacles is not part of the state space but learned implicitly during the training process.
The reward assigned to an action $\underline{a}_t$ is computed as follows: 
\begin{align}
R_{\underline{a}_t} = \alpha \cdot\frac{d_{T_{t}} - d_{T_{t+1}}}{d_{T_{init}}} + \beta \cdot P(d_{C_{t+1}}),
\end{align}
where $\alpha$ and $\beta$ are weighing factors, $d_T$ is the distance to the target point and $P$ is a quadratic penalty function depending on $d_{C_{t+1}}$, the minimum distance to a collision at $t+1$.
The first term of the reward function encourages the robot to move as fast as possible towards the target point, thus maximizing $d_{T_{t}} - d_{T_{t+1}}$, the difference between the distance to the target point before and after action $\underline{a}_t$ is executed. 
The penalty $P(d_{C_{t+1}})$ helps to avoid collisions, but does not provide safety guarantees by itself. 
All robot arms are controlled by a single neural network with 256 neurons in the first hidden layer and 128 neurons in the second hidden layer.
The neural network is trained using an RL algorithm called Proximal Policy Optimization (PPO) \cite{schulman2017proximal}.
The time interval between network decisions is set to $\Delta t_{N}=$ \SI{0.1}{\s}. We refer to our publicly available networks for further details on the hyperparameters used during the training process.
\subsection{Evaluation of our collision prevention method}
The average performance of our collision prevention method is evaluated for various experimental configurations considering random agents (behavior at the beginning of a training process) and trained agents.
The results are shown in TABLE \ref{table:collision_prevention}.
It can be seen that the average number of target points reached within one episode increases significantly during training.
For both random agents and trained agents, the closest distance between the observed obstacle-link pairs and link-link pairs is always greater than zero, showing that collisions are entirely prevented. For all of our experiments, we specify a safety distance $d_s$ of \SI{1.0}{\centi\meter}. The closest distance is slightly smaller than the safety distance for two reasons: Firstly, the closest distance is measured at a frequency of $f_S =$ \SI{240}{\hertz}, whereas the backup trajectory is checked for collisions at a rate of $f_C =$ \SI{100}{\hertz}. Secondly, the closest distance refers to actual values whereas the safety distance is defined with respect to trajectory setpoints. The resulting inaccuracy can always be compensated by increasing the safety distance $d_s$. 
The adaptation rate specifies the proportion of decision steps in which the alternative behavior is selected to prevent safety violations. 
The highest adaptation rate occurs in case of the four-armed robot ARMAR-6x4, indicating that this environment is particularly challenging. 
Interestingly, the four-armed robot uses only two of its arms if a single target point should be reached. The other two arms are stretched so that no collisions occur. We conclude that the additional arms are more of a burden than \mbox{a help for this particular task}.  
\begin{table}[t]
    \caption{%
    Ablation studies for collision prevention. 
    }
    \vspace{-0.15cm}
    \makegapedcells
\begin{tabular*}{0.49\textwidth}{p{35mm}p{8mm}p{16mm}p{14mm}} 
    \toprule
Configuration & \multicolumn{1}{c}{Target}  & \multicolumn{1}{c}{Episodes with} & Adaptation  \\
&  \multicolumn{1}{c}{points} &  \multicolumn{1}{c}{collisions} & \multicolumn{1}{c}{rate} \\
    \hline
ARMAR-6 with\newline  simultaneous target points & & &  \\

\tabitem Random agent without \newline \phantom{\tabitem }collision prevention
    & \vspace{-0.11cm} \hspace{-0.04cm}  $0.07$  & \vspace{-0.11cm} \hfil \SI{75.0}{\percent}   & \vspace{-0.11cm} \hfil --      \\

\tabitem Trained and evaluated \newline \phantom{\tabitem }without collision \newline \phantom{\tabitem }prevention
     & \vspace{0.08cm} \hspace{-0.125cm} $18.57$ &  \vspace{0.08cm} \hfil \SI{64.8}{\percent}    & \vspace{0.08cm} \hfil --    \\

\tabitem  Trained without collision \newline \phantom{\tabitem }prevention, evaluated \newline \phantom{\tabitem }with collision prevention
    & \vspace{0.08cm} \hspace{-0.125cm} $14.86$ &  \vspace{0.08cm} \hfil \SI{0.0}{\percent} & \vspace{0.085cm} \hfil \SI{23.0}{\percent}        \\
    
\tabitem  Trained and evaluated \newline \phantom{\tabitem }with collision prevention
    & \vspace{-0.11cm} \hspace{-0.125cm} $16.08$ &  \vspace{-0.11cm}  \hfil \SI{0.0}{\percent} & \vspace{-0.11cm} \hfil \SI{9.1}{\percent}   \\    
    
\tabitem  Trained and evaluated \newline \phantom{\tabitem }with collision prevention 
\newline \phantom{\tabitem }using setpoints
    & \vspace{0.08cm} \hspace{-0.125cm} $17.60$ &  \vspace{0.08cm} \hfil \SI{0.0}{\percent} & \vspace{0.08cm} \hfil \SI{7.0}{\percent}      \\  
    \bottomrule
    \end{tabular*}
    \vspace{-0.46cm}
    
\label{table:ablation_studies}
\end{table}
\mbox{To further analyze} the impact of our method, we perform the ablation studies shown in TABLE \ref{table:ablation_studies}. As expected, random agents and trained agents frequently cause collisions without collision prevention. 
It is possible to apply our method to an agent trained without collision prevention to enable safe deployment. In this case, collisions are prevented but the adaptation rate is rather high and the performance of the agent decreases. If our method is used during training and evaluation, the performance of the agent increases, as the neural network learns to avoid actions that lead to the selection of the alternative behavior. 
Since our method prevents collisions, a trajectory controller is able to accurately track the desired trajectory. 
For that reason, it is possible to use the trajectory setpoints rather than the actual values for the reward calculation. By doing so, the performance increases, as the delay caused by the trajectory controller is avoided.
In summary, preventing collisions entirely leads to a slight decrease in performance. However, the performance gap can be reduced by applying our method during training and evaluation and by utilizing that the reward for a collision-free trajectory can be assessed based on the trajectory setpoints rather than the actual values.
\subsection{Evaluation of our torque limit prevention method}
To evaluate the effectiveness of our method for preventing torque limit violations, we perform experiments with one and two industrial robots as well as with the humanoid robot ARMAR-6.
Performance metrics obtained by evaluating 900 episodes per experiment are shown in TABLE \ref{table:torque_limit}.
In a first experiment, a single industrial robot is trained with collision prevention but without torque limit prevention. Compared to our previous experiments, the maximum torque of the robot joints $\tau_{max}$ is reduced by \SI{40}{\percent}. Using this configuration, torque limit violations occur in more than half of the episodes. %
However, when evaluating the agent with both collision prevention and torque limit prevention, the torque limit violations no longer occur, while the impact on the performance is rather small. The same applies to our experiments with two industrial robots. 
As explained in section \ref{sec:background_simulation}, collisions typically lead to torque limit violations. For that reason, collisions can also be avoided by preventing torque limit violations. Our experiments with the humanoid robot ARMAR-6 show that neither a random agent nor a trained agent causes collisions or torque limit violations when evaluated with our torque limit prevention method. 
However, in contrast to our collision prevention method, no safety distance can be specified. 
For that reason, it makes sense to use the torque limit prevention in combination with our collision prevention method.

\label{sec:result}

\begin{table}[t]
    \caption{%
    Evaluation of the torque limit prevention. 
    }
    \vspace{-0.15cm}
    \makegapedcells
\begin{tabular*}{0.49\textwidth}{p{36mm}p{8mm}p{9mm}p{19mm}} 
    \toprule
Configuration &  \multicolumn{1}{c}{Target} & \multicolumn{2}{c}{Episodes with} \\
& \multicolumn{1}{c}{points} & collisions & torque violations \\
    \hline
One industrial robot, maximum torque reduced by \SI{40}{\percent} & & &  \\        
\tabitem Trained and evaluated with \newline \phantom{\tabitem }collision prevention only
    & \vspace{-0.08cm} \hspace{-0.0cm}  $8.90$  & \vspace{-0.08cm} \hfil \SI{0.0}{\percent}   & \vspace{-0.08cm} \hfil \SI{52.3}{\percent}      \\
\tabitem Trained with collision\newline \phantom{\tabitem }prevention, evaluated with \newline \phantom{\tabitem }collision prevention and\newline \phantom{\tabitem }torque limit prevention
    & \vspace{0.25cm} \hspace{-0.0cm}  $8.20$  & \vspace{0.25cm} \hfil \SI{0.0}{\percent}   & \vspace{0.25cm} \hfil \SI{0.0}{\percent}      \\
Two industrial robot,  alternating target points, maximum torque reduced by \SI{40}{\percent} & & &  \\        
\tabitem Trained and evaluated with \newline \phantom{\tabitem }collision prevention only
    & \vspace{-0.08cm} \hspace{-0.0cm}  $8.43$  & \vspace{-0.08cm} \hfil \SI{0.0}{\percent}   & \vspace{-0.08cm} \hfil \SI{84.3}{\percent}      \\
\tabitem Trained with collision\newline \phantom{\tabitem }prevention, evaluated with \newline \phantom{\tabitem }collision prevention and\newline \phantom{\tabitem }torque limit prevention
    & \vspace{0.25cm} \hspace{-0.0cm}  $6.18$  & \vspace{0.25cm} \hfil \SI{0.0}{\percent}   & \vspace{0.25cm} \hfil \SI{0.0}{\percent}      \\
\hline
ARMAR-6 with  \newline simultaneous target points & & &  \\

\tabitem Random agent with\newline \phantom{\tabitem }torque limit prevention only
    & \vspace{-0.11cm} \hspace{-0.0cm}  $0.07$  & \vspace{-0.11cm} \hfil \SI{0.0}{\percent}   & \vspace{-0.11cm} \hfil \SI{0.0}{\percent}     \\

\tabitem Trained and evaluated \newline \phantom{\tabitem }with torque limit prevention
     & \vspace{-0.11cm} \hspace{-0.1cm} $16.17$ &  \vspace{-0.11cm} \hfil \SI{0.0}{\percent}    & \vspace{-0.11cm} \hfil \SI{0.0}{\percent}    \\
     
    \bottomrule
    \end{tabular*}
    \vspace{-0.6cm}
    
\label{table:torque_limit}
\end{table}
\begin{figure}[h]
\centering
    \begin{subfigure}[c]{0.22\textwidth}
        \begin{tikzpicture}
            \node[anchor=south west,inner sep=0] (image) at (0.0cm,0) {\includegraphics[trim=370 0 188 70, clip, height=1.5\textwidth]{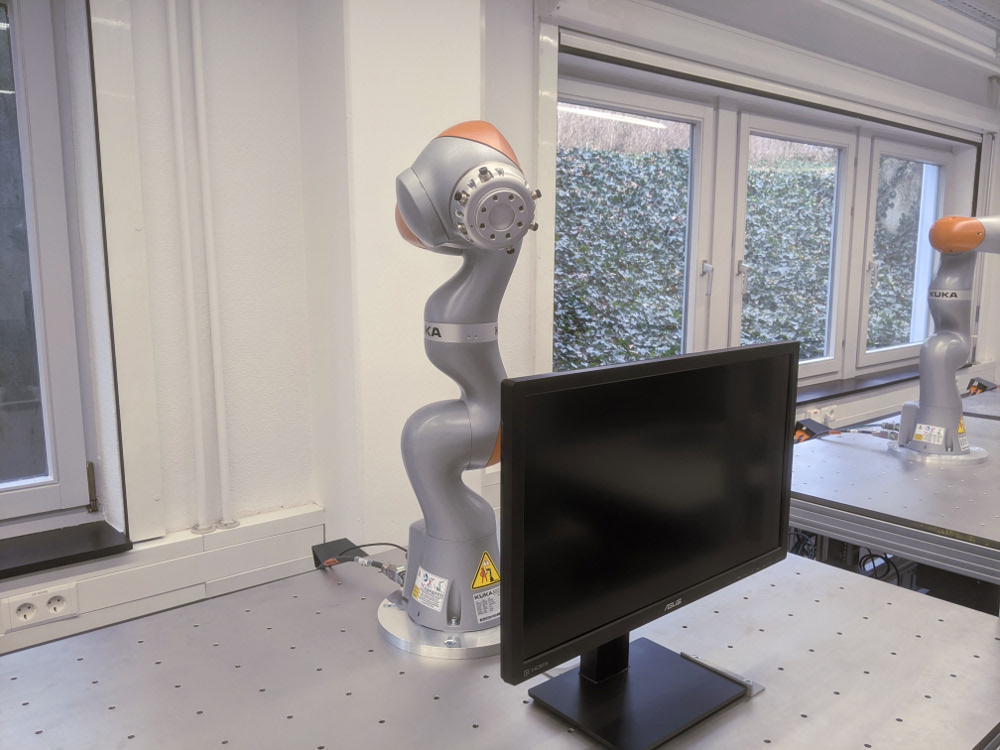}};
            \begin{scope}[x={(image.south east)},y={(image.north west)}]
                \draw[color=PATH_OKAY, line width=0.75mm, ->, >=stealth] (0.37, 0.73) arc (220:290:0.25 and 0.1);
            \end{scope}
            \node(target_point) at (2.90, 4.15) {\includegraphics[trim=0 0 0 0, clip, width=0.14\textwidth]{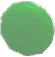}};
        \end{tikzpicture}
	   \subcaption{The target point is reached.}
	   \label{fig:real_robot_obstacle_a}
		
	\end{subfigure}
	\hspace{0.005\textwidth}
	\begin{subfigure}[c]{0.22\textwidth}
	    \begin{tikzpicture}
            \node[anchor=south west,inner sep=0] (image) at (0,0) {\includegraphics[trim=255 30 300 30, clip, height=1.5\textwidth]{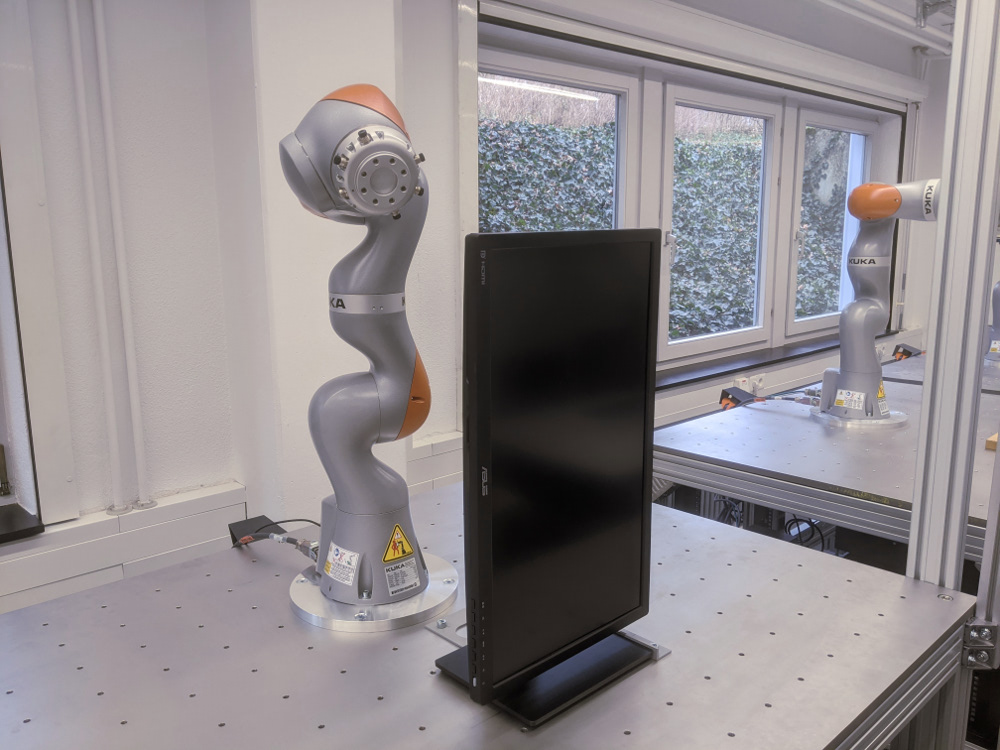}};
            \begin{scope}[x={(image.south east)},y={(image.north west)}]
                \draw[color=PATH_COLLISION, line width=0.75mm, ->, >=stealth] (0.43, 0.73) arc (220:290:0.25 and 0.1);
            \end{scope}
            \node(target_point) at (3.16, 4.15) {\includegraphics[trim=0 0 0 0, clip, width=0.14\textwidth]{figures/target_point_transparent_1.png}};
        \end{tikzpicture}
		\subcaption{The robot stops.}%
		\label{fig:real_robot_obstacle_b}

	\end{subfigure} 
	\caption{Collisions are prevented even if the shape of an obstacle is changed from (a) to (b) after training.
	} %
	\label{fig:real_robot_obstacle}
\end{figure}
\subsection{Real-time capability and sim-2-real transfer}
In order to apply our approach to real robots, all computations must be performed in real-time. 
As a first step, we analyze the maximum computation time of simulated episodes with a duration of \SI{8}{\second} using an Intel i9-9900K CPU. The results are shown in TABLE \ref{table:computation_times}.%
Using the precalculation technique described in section \ref{sec:real_time}, real-time capability is achieved if the time required for the action generation and the background simulation never \mbox{exceeds $\Delta t_N$.}
With \mbox{$\Delta t_N=$ \SI{0.1}{\second}}, this condition is met for all of our experiments apart from those involving collision prevention with ARMAR6-x4.
\begin{table}[t]
    \caption{Maximum computation times for episodes with a duration of 8 seconds. Green values indicate real-time capability. Results based on 100 episodes.}
    \vspace{-0.15cm}
    \makegapedcells
\begin{tabular*}{0.495\textwidth}{@{}p{33.5mm}p{11mm}p{11mm}p{11mm}} 
    \toprule

\hspace{0.02cm} Agent trained with    & \multicolumn{1}{c}{Industrial} & \multicolumn{2}{c}{Humanoids}  \\ %
\hspace{0.02cm} a single target point   & \multicolumn{1}{c}{One robot} & \multicolumn{1}{c}{ARMAR-6} & \multicolumn{1}{c}{ARMAR-6x4} \\
    \hline
  \tabitem Neither collision \newline\phantom{\tabitem}prevention nor torque \newline\phantom{\tabitem}limit prevention
    & \vspace{0.08cm} \hfil \textcolor{TABLE_GOOD}{\SI{0.57}{\second}}  &  \vspace{0.08cm} \hfil \textcolor{TABLE_GOOD}{\SI{0.65}{\second}}      & \vspace{0.08cm} \hfill \textcolor{TABLE_GOOD}{\SI{0.70}{\second}}   \\
 \tabitem Collision prevention 
      & \hfil \textcolor{TABLE_GOOD}{\SI{0.96}{\second}}  & \hfil \textcolor{TABLE_GOOD}{\SI{2.43}{\second}}      &  \hfill \textcolor{black}{\SI{7.82}{\second}}    \\
   \tabitem Torque limit prevention
      & \hfil \textcolor{TABLE_GOOD}{\SI{1.31}{\second}}  & \hfil \textcolor{TABLE_GOOD}{\SI{2.63}{\second}}      &  \hfill \textcolor{TABLE_GOOD}{\SI{6.10}{\second}}    \\   
   \tabitem Collision prevention and  \newline\phantom{\tabitem}torque limit prevention
      & \vspace{-0.05cm} \hfil \textcolor{TABLE_GOOD}{\SI{1.59}{\second}}  & \vspace{-0.05cm} \hfil \textcolor{TABLE_GOOD}{\SI{4.23}{\second}}      & \vspace{-0.05cm}\hspace*{0.25cm} \hfill  \SI{12.88}{\second}  \\ 

    \bottomrule
    \end{tabular*}
    \vspace{-0.56cm}
\label{table:computation_times}
\end{table}
We note that the computational effort for collision prevention can be significantly reduced by decreasing the rate of collision checks $f_C$, which is set to \SI{100}{\hertz} in our experiments. When doing so, the safety distance $d_s$ should be increased to ensure that no collisions occur between the collision checks.
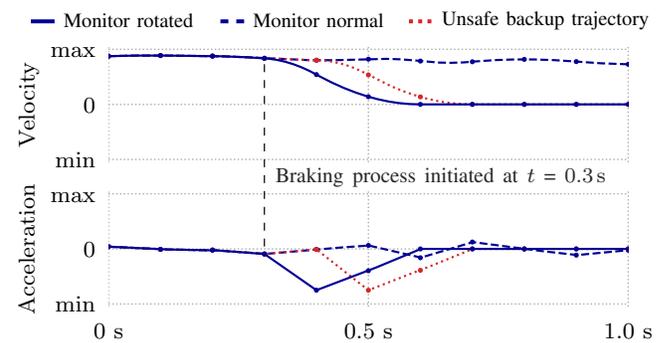
\begin{figure}[b]
\vspace{-0.5cm}
    \begin{tikzpicture}
                \def\xminPlot{0.151} 
                \def\xmaxPlot{0.949} 
                \SUBTRACT{\xmaxPlot}{\xminPlot}{\xdeltaPlot}
                \DIVIDE{\xdeltaPlot}{3}{\xdeltaPlotNorm}
                \def\yminPlot{0.135}
                \def\ymaxPlot{0.985}
                \def\yminAcc{0.364}
                \def\ymaxAcc{0.535}
                \SUBTRACT{\ymaxPlot}{\yminPlot}{\ydeltaPlot}
                \SUBTRACT{\ymaxAcc}{\yminAcc}{\ydeltaAcc}
                \MULTIPLY{\ydeltaAcc}{4}{\ydeltaGraph}
                \SUBTRACT{\ydeltaPlot}{\ydeltaGraph}{\ydeltaTmp}
                \DIVIDE{\ydeltaTmp}{3}{\ydeltaGap}
                
                 \node[text width=4cm] at (0.2cm, 4.35cm) (origin){};
                	\draw [draw=LINE_COLOR_BLUE, line width=1.5pt, scale=0.7] ($(origin.center)+(-5.9cm, -0.0cm)$) -- + (0.45cm, 0cm) node[pos=1, right, yshift=0.01cm, align=left, scale=0.85]{\small{Monitor rotated}};
            	    \draw [dashed, draw=LINE_COLOR_BLUE, line width=1.5pt, scale=0.7] ($(origin.center)+(-2.3cm, 0.0cm)$) -- + (0.45cm, 0cm) node[pos=1, right, yshift=0.01cm, align=left, scale=0.85]{\small{Monitor normal}};
            	    \draw [dotted, draw=LINE_COLOR_RED, line width=1.5pt, scale=0.7] ($(origin.center)+(1.28cm, -0.0cm)$) -- + (0.45cm, 0cm) node[pos=1, right, yshift=0.01cm, align=left, scale=0.85]{\small{Unsafe backup trajectory}};
            	\node[anchor=south west,inner sep=0] (image) at (-4.2cm,0) {\includegraphics[trim=0 0 0 0, clip, width=\linewidth]{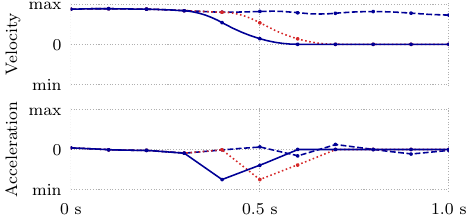}};
                   \draw[color=black!90, line width=0.175mm, dashed] (-0.826, 3.78) -- (-0.826, 1.24);
                    \node[color=black!85, rectangle, anchor=west, text width=6cm] (target_point_text) at (-0.8, 2.28) {\footnotesize{Braking process initiated at $t$ = \SI{0.3}{\second}}};
    \end{tikzpicture}
	\vspace{-0.6cm}
	\caption{
	 Joint trajectories generated by the same neural network for the two monitor orientations  shown \mbox{in Fig. \ref{fig:real_robot_obstacle}.}}
	\label{fig:real_robot_trajectories}
\end{figure}

As can be seen in the accompanying video, we successfully transferred a neural network trained with both collision prevention and torque limit prevention to a real industrial robot shown in Fig \ref{fig:real_robot_obstacle}.
For real-time execution, two parallel threads are started by our program: 
At time step $t$, the first thread starts to perform the action generation and the background simulation for time step $t+1$ leading to a safely executable joint acceleration $a_{t+2}$. %
The second thread keeps synchronized with the position-based trajectory controller of the KUKA iiwa robot
and uses the computed joint acceleration to interpolate position setpoints at a \mbox{frequency of \SI{200}{\hertz}}.
To demonstrate that collisions are prevented even if the environment is changed after training, we rotate the monitor in front of the robot so that a desired target point can no longer be reached (Fig. \ref{fig:real_robot_obstacle_b}). For both monitor orientations, the resulting acceleration and velocity of an exemplary robot joint is visualized in Fig. \ref{fig:real_robot_trajectories}.
In case of the rotated monitor, a braking process is initiated at $t$ = \SI{0.3}{\second} to prevent a collision between the robot and the monitor.

\section{Discussion}
Our method succeeded in learning online generation of optimized robot motions without causing collisions and torque limit violations.
In addition, our action space allows the robot joints to be safely operated at their kinematic limits, which is crucial for learning fast and dynamic robot motions like the ones shown in our video. 
Compared to model predictive control, our method offers the advantage that no dynamics model is needed and that the reward function does not need to be differentiable with respect to the actions. As a drawback, a training phase is needed to generate optimized motions. 
By adjusting the task-specific part of the state and the definition of the reward function, our method can also be used to learn tasks other than the ones shown in this work. 
In addition, it is possible to generate a dataset of random but collision-free trajectories, which can be used to learn time-optimized path tracking via reinforcement learning \cite{kiemel2022path}. 
While our approach provides explicit safety guarantees, a current limitation is that collisions with arbitrarily moving obstacles
are not explicitly considered. For most industrial environments, it is reasonable to assume that the position and shape of each obstacle are known before training. %
However, computing an optimized backup trajectory that actively avoids moving obstacles is a promising direction to address this limitation in future work. 

\section{Conclusion and future work}
\label{sec:conclusion}

This paper presented a method to learn collision-free and torque-limited robot trajectories. 
Conflicting constraints are avoided by ensuring the existence of an alternative safe behavior at each decision step.
Violations of kinematic joint limits are prevented by the design of the action space used for reinforcement learning.  
Our evaluation showed that neither random agents nor trained agents violated the specified constraints. 
Experiments with a real industrial robot demonstrated that safe motions can be computed in real-time. 

In future work, we would like to apply our approach to other safety constraints like Cartesian velocity limits.   
In addition, we are interested in investigating optimized backup trajectories that actively avoid moving obstacles. %

\section*{ACKNOWLEDGMENT}

This research was supported by the German Federal Ministry of Education and Research (BMBF) and the Indo-German Science \& Technology Centre (IGSTC) as part of the project TransLearn (01DQ19007A). 
We thank Tamim Asfour for his valuable feedback and advice.

\bibliographystyle{IEEEtran}
\bibliography{root}

\end{document}

%% file: figures/asb_demo.tex
\newcommand{\addImage}[2]{
\def\imagewidth{0.125}\ifnum #1 = 0
\includegraphics[trim=185 200 1035 240, clip, width=\imagewidth\textwidth]{#2}\else
\includegraphics[trim=1040 200 180 240, clip, width=\imagewidth\textwidth]{#2}\fi}

\newcommand{\timeStepImageSingle}[5]{
\begin{tikzpicture}[scale=1.0, every node/.style={scale=1}]
\ifnum #4 = 0
        \node[draw, rectangle, draw=#5](last_image) at (0, 0) {\addImage{#1}{#2}};
      \else
        \node[draw, rectangle, draw=#5, very thick](last_image) at (0, 0) {\addImage{#1}{#2}};
\fi
\node[left of=last_image, node distance=0.67cm, yshift=0.7cm, scale=0.85]{\textcolor{black}{#3}};
\end{tikzpicture}}

\newcommand{\timeStepImageSeries}[6]{

\begin{tikzpicture}[scale=1.0, every node/.style={scale=0.8}]
\begin{scope}[cm={1,0.3,0,1,(1,0)}]
     \node[transform shape, draw, inner sep=0.0mm](image_1) {\addImage{#1}{#2}};
\end{scope}
\begin{scope}[cm={1,0.3,0,1,(1.1,0)}]
      \node[transform shape, draw, inner sep=0.0mm](image_2) {\addImage{#1}{#2}};
\end{scope}
\begin{scope}[cm={1,0.3,0,1,(1.2,0)}]
      \ifnum #4 = 0
        \node[transform shape, draw=#5, inner sep=0.0mm](image_3) {\addImage{#1}{#2}};
      \else
        \node[transform shape, draw=#5, very thick, inner sep=0.0mm](image_3) {\addImage{#1}{#2}};
      \fi
      \node[transform shape, left of=image_3, node distance=0.62cm, yshift=0.7cm, scale=0.825]{\textcolor{black!80}{#3}};
      \node[transform shape, right of=image_3, node distance=0.62cm, yshift=0.7cm, scale=0.825]{\textcolor{black!80}{#6}};
\end{scope}
\end{tikzpicture} 
}

\def\neuronSize{0.15cm}
\def\neuronInterLayerDistance{0.275cm}
\def\neuronLayerDistance{0.55cm}
\tikzstyle{first_layer} = [draw={black}, fill={LINE_COLOR_BLUE!15}, circle, minimum size=\neuronSize, inner sep=0pt, outer sep=0pt]
\tikzstyle{middle_layer} = [draw=black, fill=black!5, circle, minimum size=\neuronSize, inner sep=0pt, outer sep=0pt]
\tikzstyle{last_layer} = [draw={black}, fill={LINE_COLOR_RED!15}, circle, minimum size=\neuronSize, inner sep=0pt, outer sep=0pt]
\tikzstyle{neuron_connection} = [draw=black!80]
\newcommand{\addNeuralNetwork}{
\begin{tikzpicture}[scale=1.0, every node/.style={scale=1}]
\node[middle_layer](middle_0) at (0, 0) {};
\node[middle_layer, above of=middle_0, node distance=\neuronInterLayerDistance](middle_t_1){};
\node[middle_layer, above of=middle_t_1, node distance=\neuronInterLayerDistance](middle_t_2){};
\node[middle_layer, below of=middle_0, node distance=\neuronInterLayerDistance](middle_b_1){};
\node[middle_layer, below of=middle_b_1, node distance=\neuronInterLayerDistance](middle_b_2){};
\node[last_layer, right of=middle_0, node distance=\neuronLayerDistance](last_0){};
\node[last_layer, above of=last_0, node distance=\neuronInterLayerDistance](last_t_1){};
\node[last_layer, below of=last_0, node distance=\neuronInterLayerDistance](last_b_1){};
\node[above of=middle_0, node distance=0.5*\neuronInterLayerDistance](middle_t_helper){};
\node[first_layer, left of=middle_t_helper, node distance=\neuronLayerDistance](first_t_1){};
\node[first_layer, above of=first_t_1, node distance=\neuronInterLayerDistance](first_t_2){};
\node[first_layer, below of=first_t_1, node distance=\neuronInterLayerDistance](first_b_1){};
\node[first_layer, below of=first_b_1, node distance=\neuronInterLayerDistance](first_b_2){};
\draw [neuron_connection] (first_t_1.0) -- (middle_t_1.180);
\draw [neuron_connection] (first_t_1.0) -- (middle_t_2.180);
\draw [neuron_connection] (first_t_1.0) -- (middle_0.180);
\draw [neuron_connection] (first_t_1.0) -- (middle_b_1.180);
\draw [neuron_connection] (first_t_1.0) -- (middle_b_2.180);

\draw [neuron_connection] (first_t_2.0) -- (middle_t_1.180);
\draw [neuron_connection] (first_t_2.0) -- (middle_t_2.180);
\draw [neuron_connection] (first_t_2.0) -- (middle_0.180);
\draw [neuron_connection] (first_t_2.0) -- (middle_b_1.180);
\draw [neuron_connection] (first_t_2.0) -- (middle_b_2.180);

\draw [neuron_connection] (first_b_1.0) -- (middle_t_1.180);
\draw [neuron_connection] (first_b_1.0) -- (middle_t_2.180);
\draw [neuron_connection] (first_b_1.0) -- (middle_0.180);
\draw [neuron_connection] (first_b_1.0) -- (middle_b_1.180);
\draw [neuron_connection] (first_b_1.0) -- (middle_b_2.180);

\draw [neuron_connection] (first_b_2.0) -- (middle_t_1.180);
\draw [neuron_connection] (first_b_2.0) -- (middle_t_2.180);
\draw [neuron_connection] (first_b_2.0) -- (middle_0.180);
\draw [neuron_connection] (first_b_2.0) -- (middle_b_1.180);
\draw [neuron_connection] (first_b_2.0) -- (middle_b_2.180);

\draw [neuron_connection] (middle_0.0) -- (last_0.180);
\draw [neuron_connection] (middle_0.0) -- (last_t_1.180);
\draw [neuron_connection] (middle_0.0) -- (last_b_1.180);

\draw [neuron_connection] (middle_t_1.0) -- (last_0.180);
\draw [neuron_connection] (middle_t_1.0) -- (last_t_1.180);
\draw [neuron_connection] (middle_t_1.0) -- (last_b_1.180);

\draw [neuron_connection] (middle_t_2.0) -- (last_0.180);
\draw [neuron_connection] (middle_t_2.0) -- (last_t_1.180);
\draw [neuron_connection] (middle_t_2.0) -- (last_b_1.180);

\draw [neuron_connection] (middle_b_1.0) -- (last_0.180);
\draw [neuron_connection] (middle_b_1.0) -- (last_t_1.180);
\draw [neuron_connection] (middle_b_1.0) -- (last_b_1.180);

\draw [neuron_connection] (middle_b_2.0) -- (last_0.180);
\draw [neuron_connection] (middle_b_2.0) -- (last_t_1.180);
\draw [neuron_connection] (middle_b_2.0) -- (last_b_1.180);
\end{tikzpicture}}

\tikzstyle{image_collection} = [inner sep=0pt, %
    minimum height=0cm, minimum width=0cm, align=center]
\begin{tikzpicture}[scale=1.0, every node/.style={scale=1.0}]
    \tikzstyle{arrow_style} = [->, draw=black]
    \def\distanceTopBottom{3.0cm}
    \def\colorFinalFirst{LINE_COLOR_BLUE}
    \def\colorFinalSecond{cyan}
    \def\colorCollision{LINE_COLOR_RED}
    \def\colorAnnotations{black!70}
    \node[outer sep=4.0pt, image_collection](t_0_start) at (0cm, 0cm) {\timeStepImageSingle{0}{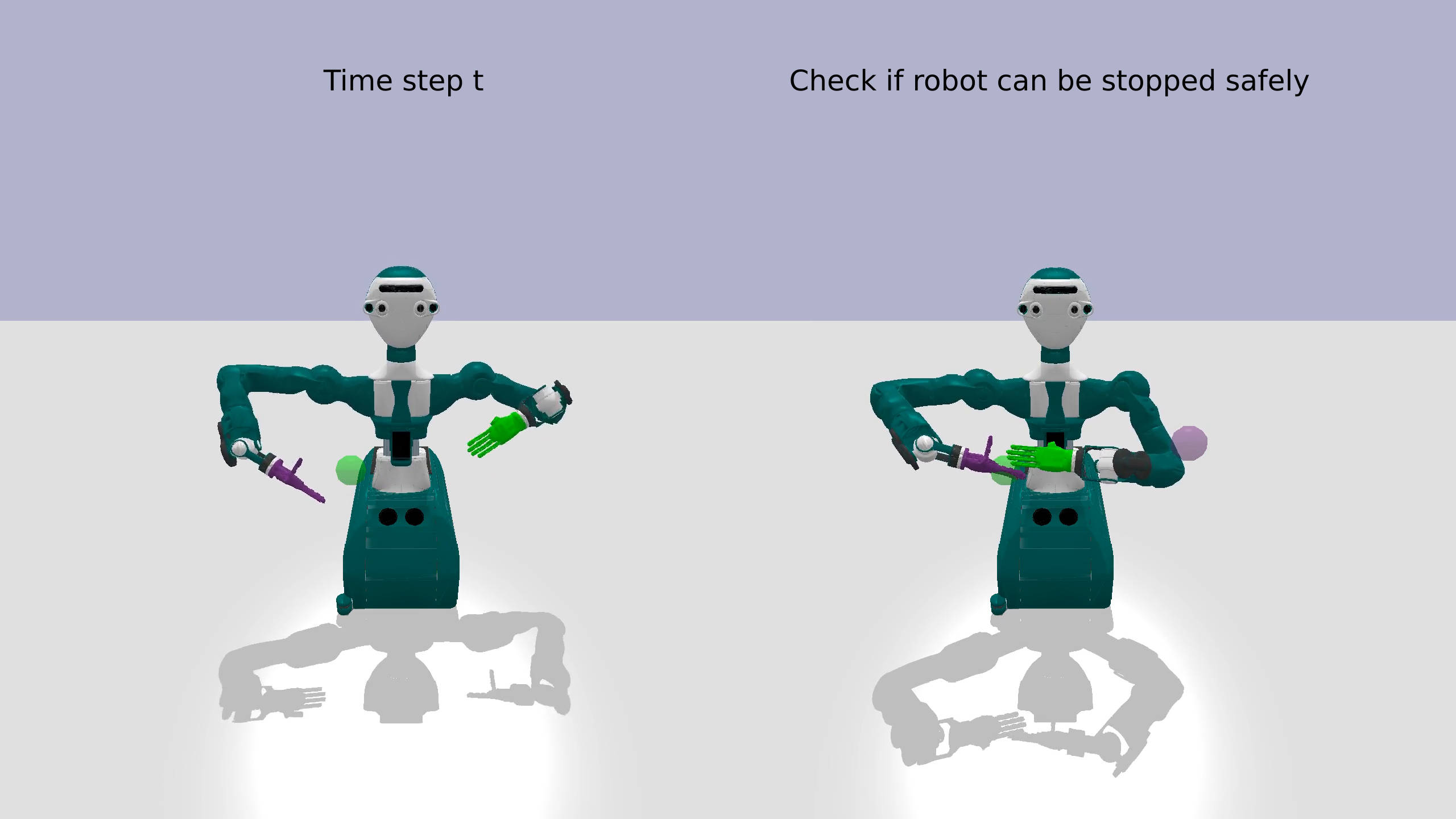}{$a_{t}$}{0}{black}};
    \node[left of=t_0_start, node distance=1.5cm, scale=0.9](t_0_start_text){\rot{Time step $t$}};
    
    \node[image_collection, outer sep=4.0pt, below of=t_0_start, node distance=\distanceTopBottom](t_1_start)
    {\timeStepImageSingle{0}{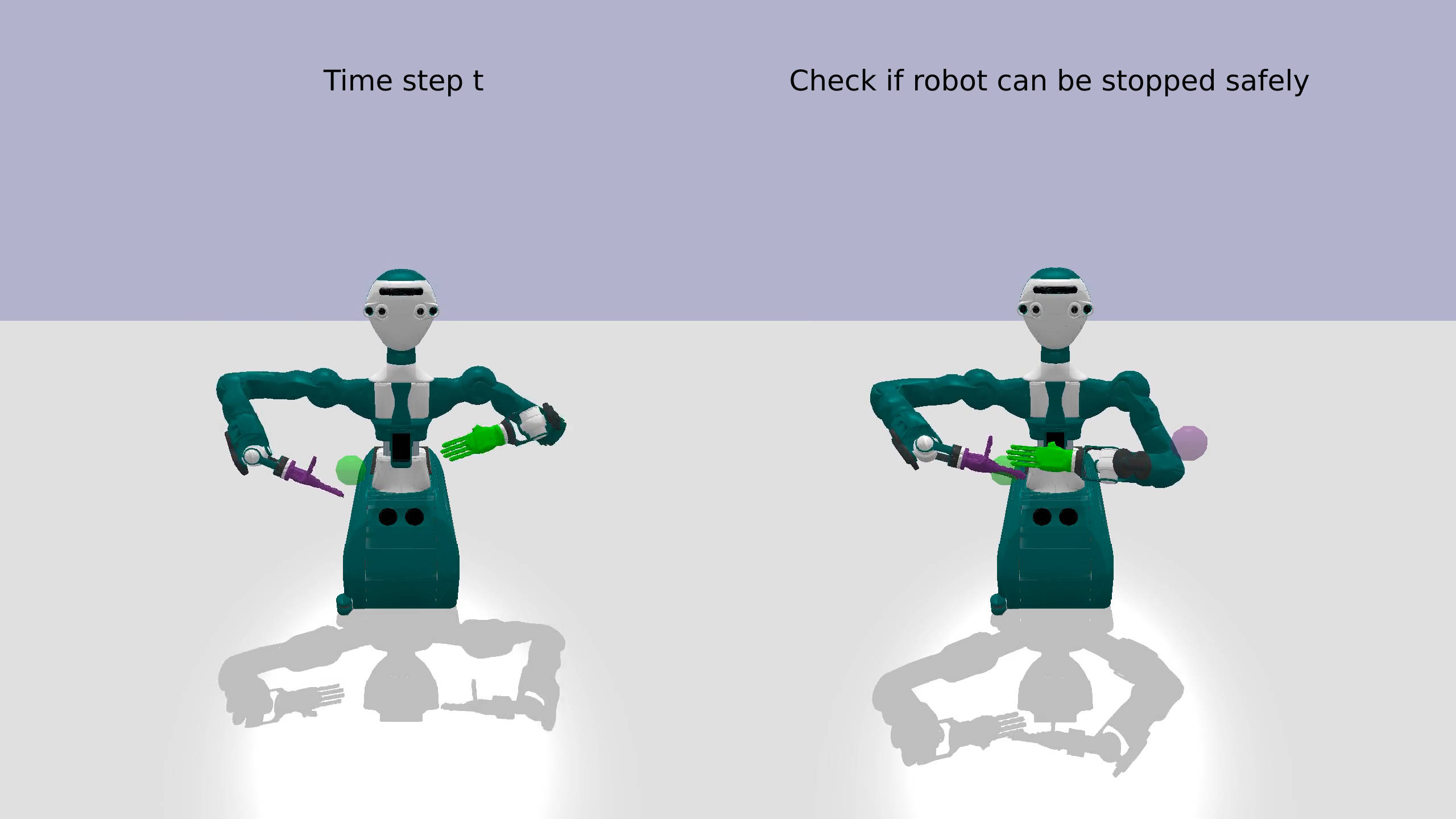}{$a_{t+1}$}{1}{\colorFinalFirst}};
    \node[left of=t_1_start, node distance=1.5cm, scale=0.9](t_1_start_text){\rot{Time step $t+1$}};
    
    \node[right of=t_0_start, node distance=2.75cm, minimum width=0.0cm, inner sep=3pt] (network_top){\addNeuralNetwork};
    \draw [{arrow_style}] (t_0_start.0) -- node[pos=0.5, above]{$s_{t}$} (network_top.180);
    
    \node[right of=network_top, node distance=2.75cm, outer sep=2.0pt, yshift=0.0cm,  image_collection](t_0_action){\timeStepImageSeries{0}{figures/asb_t_0/left_t_end_right_t_brake_end_small.jpg}{$a_{t+1_N}$}{1}{\colorFinalFirst}{}};
    \draw [{arrow_style}] (network_top.0) -- (t_0_action.180) node[pos=0.5, above=-0.04cm]{$\underline{a}_{t}$};

    \node[right of=t_0_action, node distance=2.25cm, outer sep=1.5pt, yshift=0.0cm,  image_collection](t_0_brake_1){\timeStepImageSeries{1}{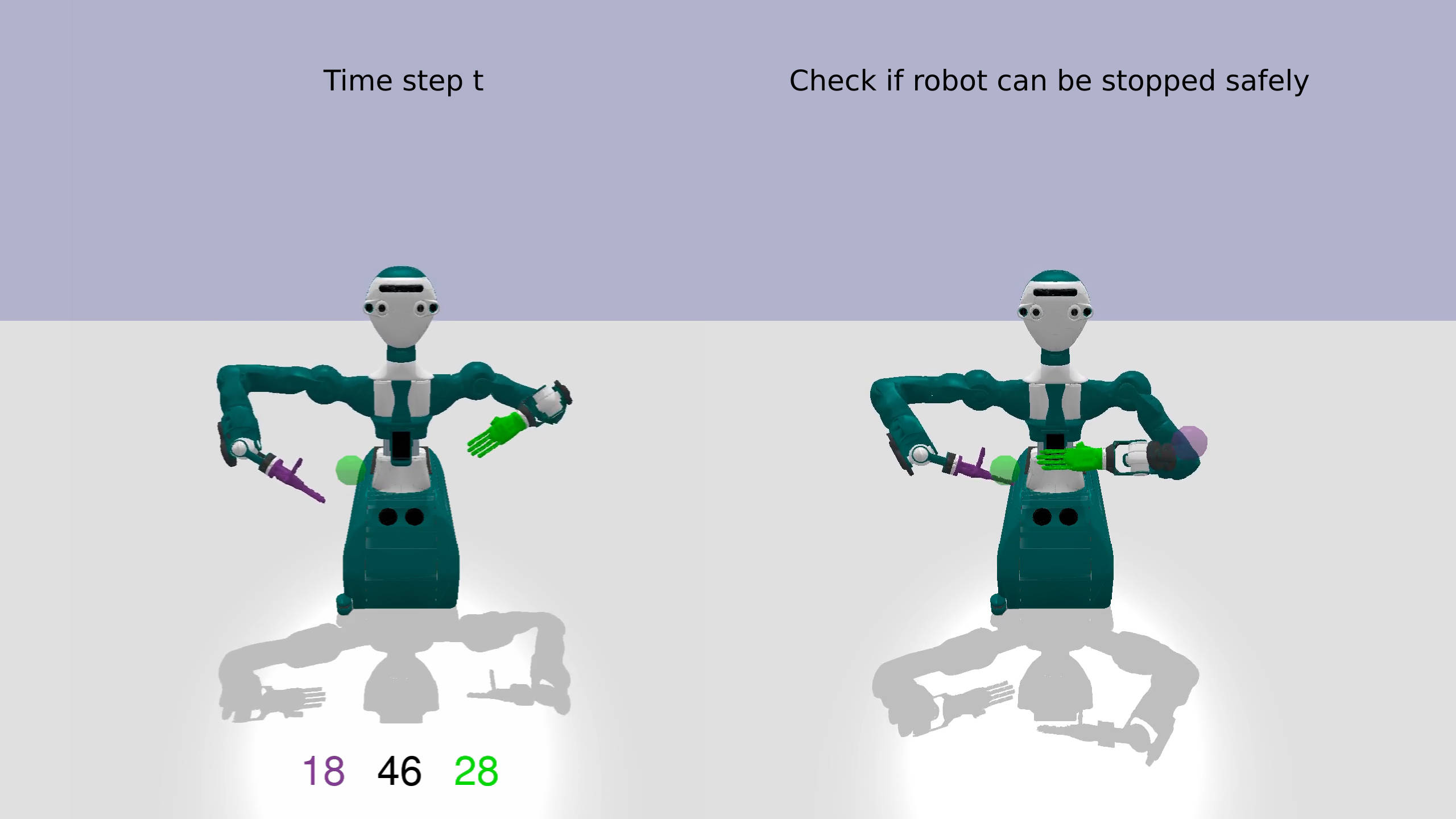}{$a_{t+2_B}'$}{1}{\colorFinalSecond}{}};
    \node[right of=t_0_brake_1, node distance=1.3cm, outer sep=1.5pt, yshift=0.0cm,  image_collection](t_0_brake_2){\timeStepImageSeries{1}{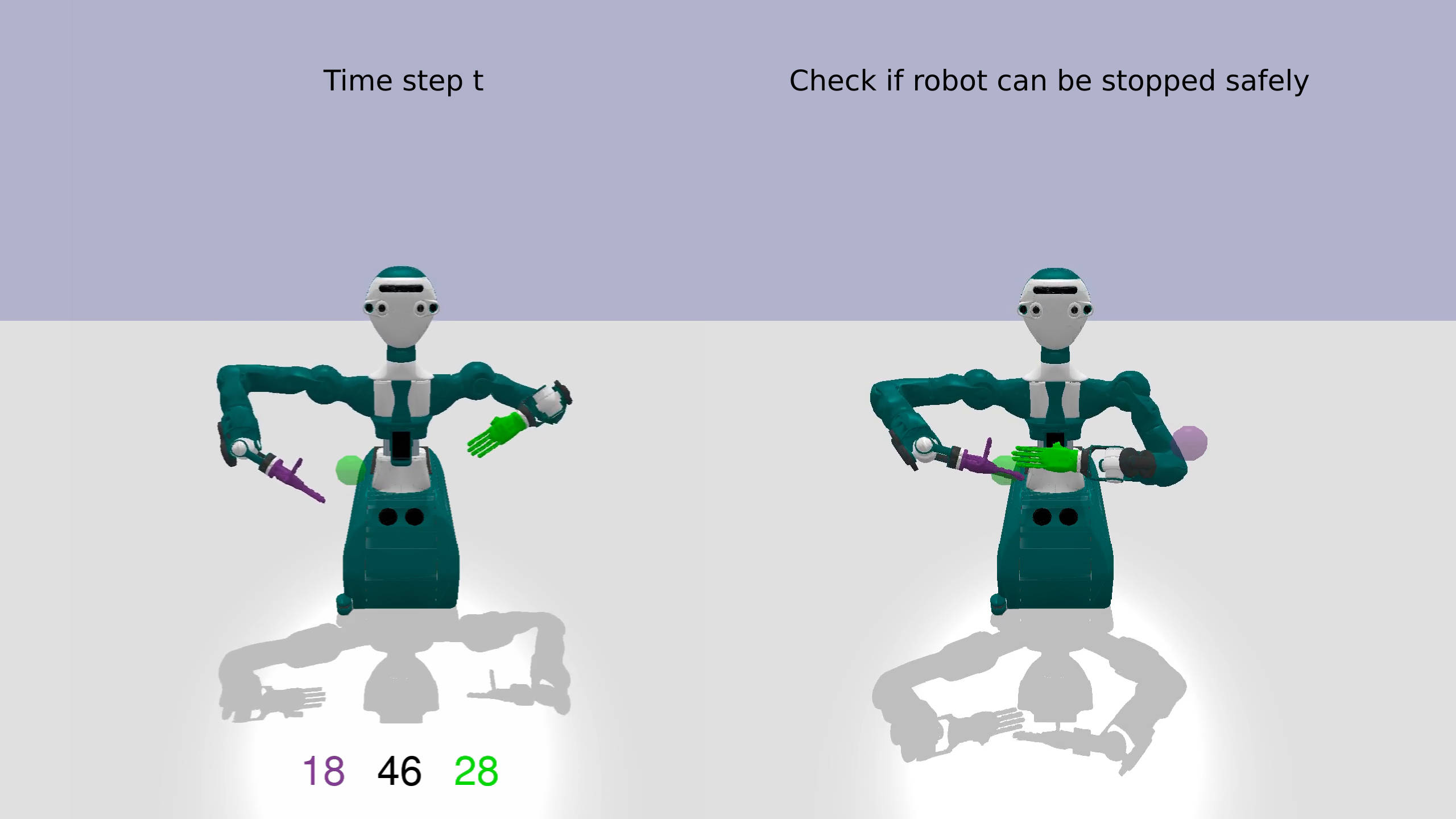}{$a_{t+3_B}'$}{0}{black}{}};
    \node[right of=t_0_brake_2, node distance=1.3cm, outer sep=1.5pt, yshift=0.0cm,  image_collection](t_0_brake_3){\timeStepImageSeries{1}{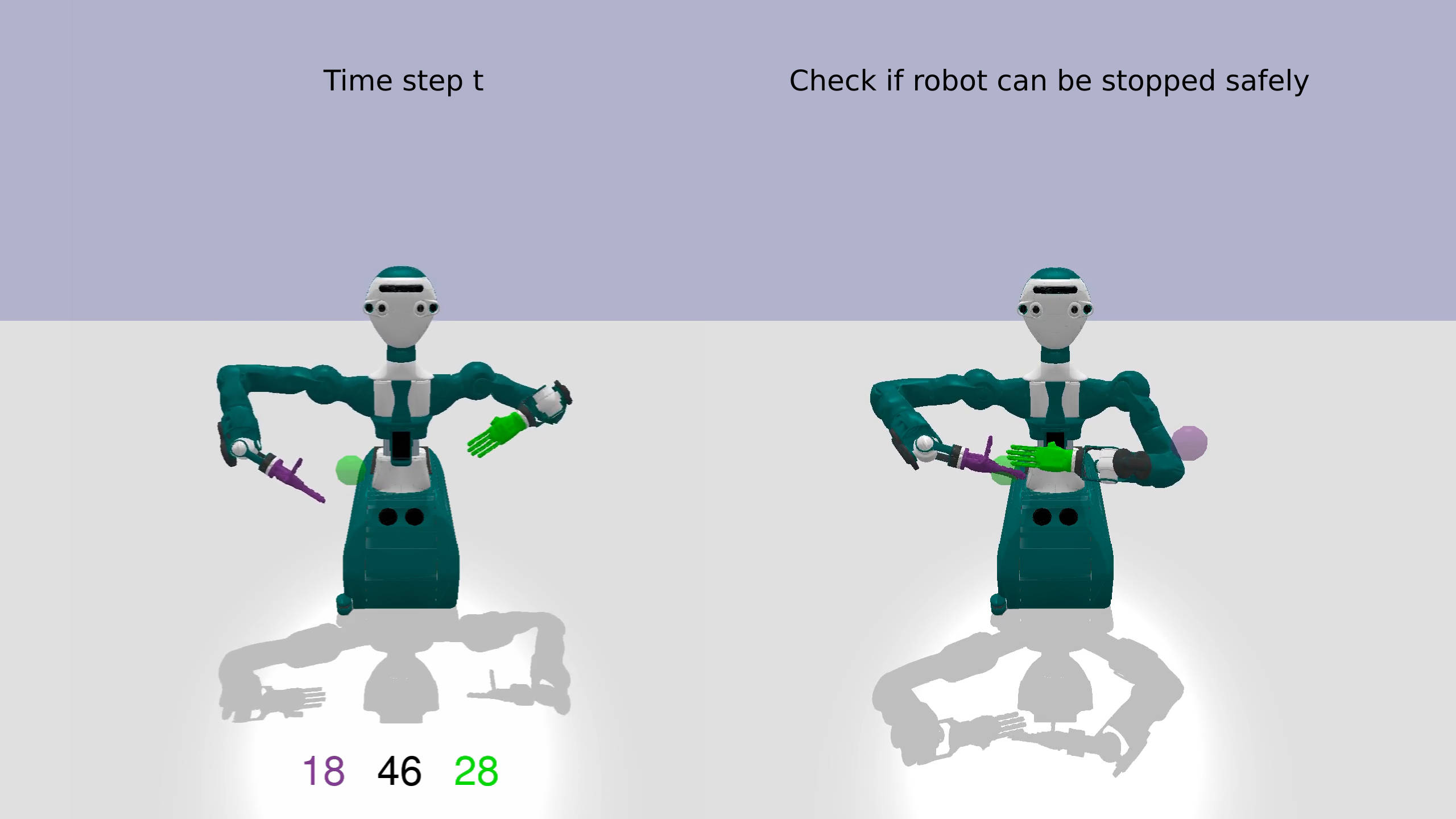}{$a_{t+4_B}'$}{0}{black}{}};
    
    \node[right of=t_0_brake_3, node distance=4.3cm, outer sep=4.0pt, yshift=0.0cm,  image_collection](t_0_final){\timeStepImageSeries{0}{figures/asb_t_0/left_t_end_right_t_brake_end_small.jpg}{$a_{t+1}$}{1}{\colorFinalFirst}{$a_{t+1_N}$}};
    
    \path [] (t_0_brake_3.0) -- (t_0_final.180) node[pos=0.5, above=-0.175cm, scale=0.8, align=center]{No safety \\ violations:} node[pos=0.5, below=0.15cm, scale=0.85]{$a_{t+1} = a_{t+1_N}$};
    
    \node[below of=network_top, node distance=\distanceTopBottom, minimum width=0.0cm, inner sep=3pt] (network_bottom){\addNeuralNetwork};
    \draw [{arrow_style}] (t_1_start.0) -- node[pos=0.45, above]{$s_{t+1}$} (network_bottom.180);
    
    \node[below of=t_0_action, node distance=\distanceTopBottom, outer sep=2.0pt, yshift=0.0cm,  image_collection](t_1_action){\timeStepImageSeries{1}{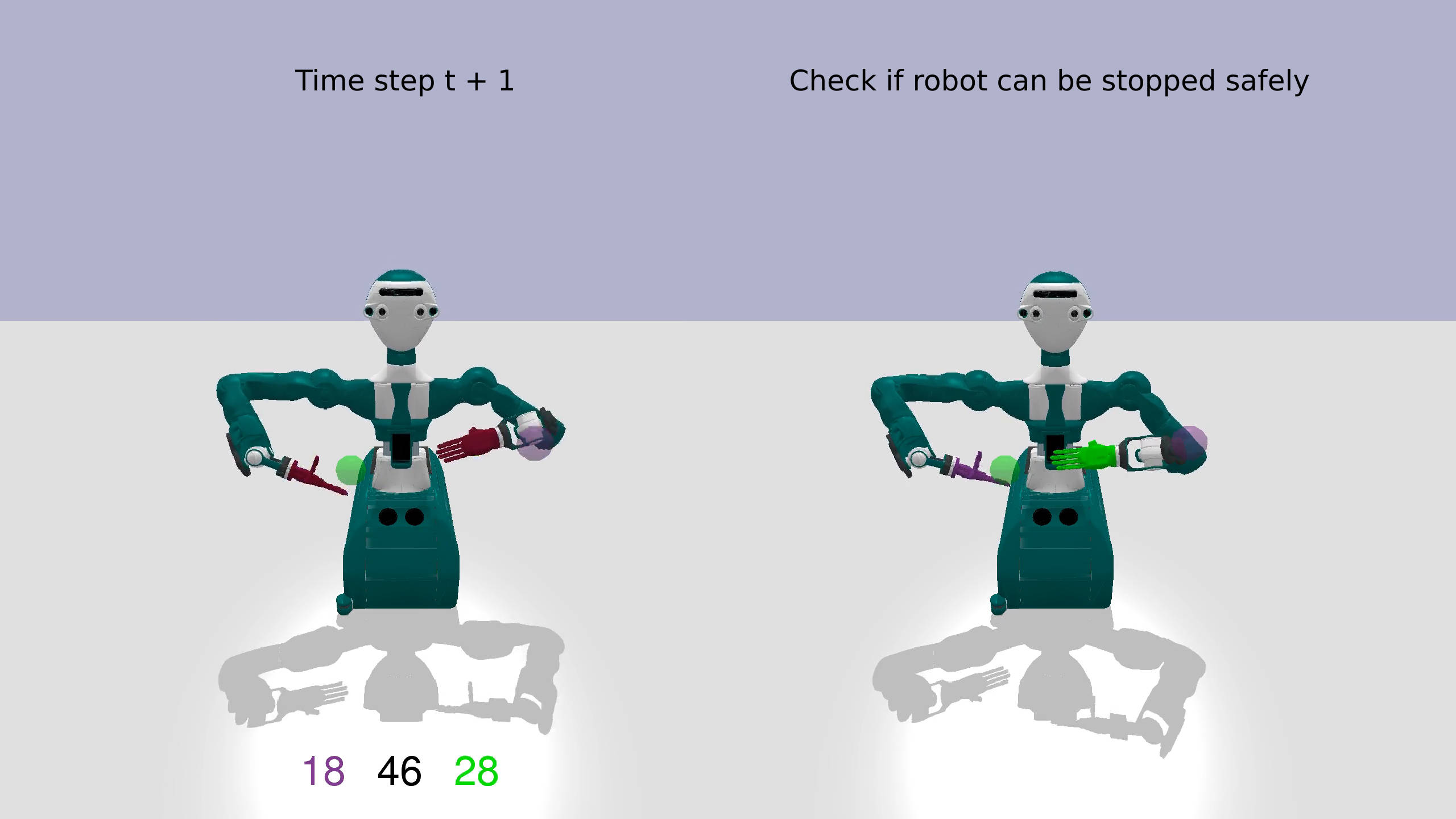}{$a_{t+2_N}$}{0}{black}{}};
    \draw [{arrow_style}] (network_bottom.0) -- (t_1_action.180) node[pos=0.5, above=-0.04cm]{$\underline{a}_{t+1}$};
    
    \node[below of=t_0_brake_1, node distance=\distanceTopBottom, outer sep=1.5pt, yshift=0.0cm,  image_collection](t_1_brake_1){\timeStepImageSeries{1}{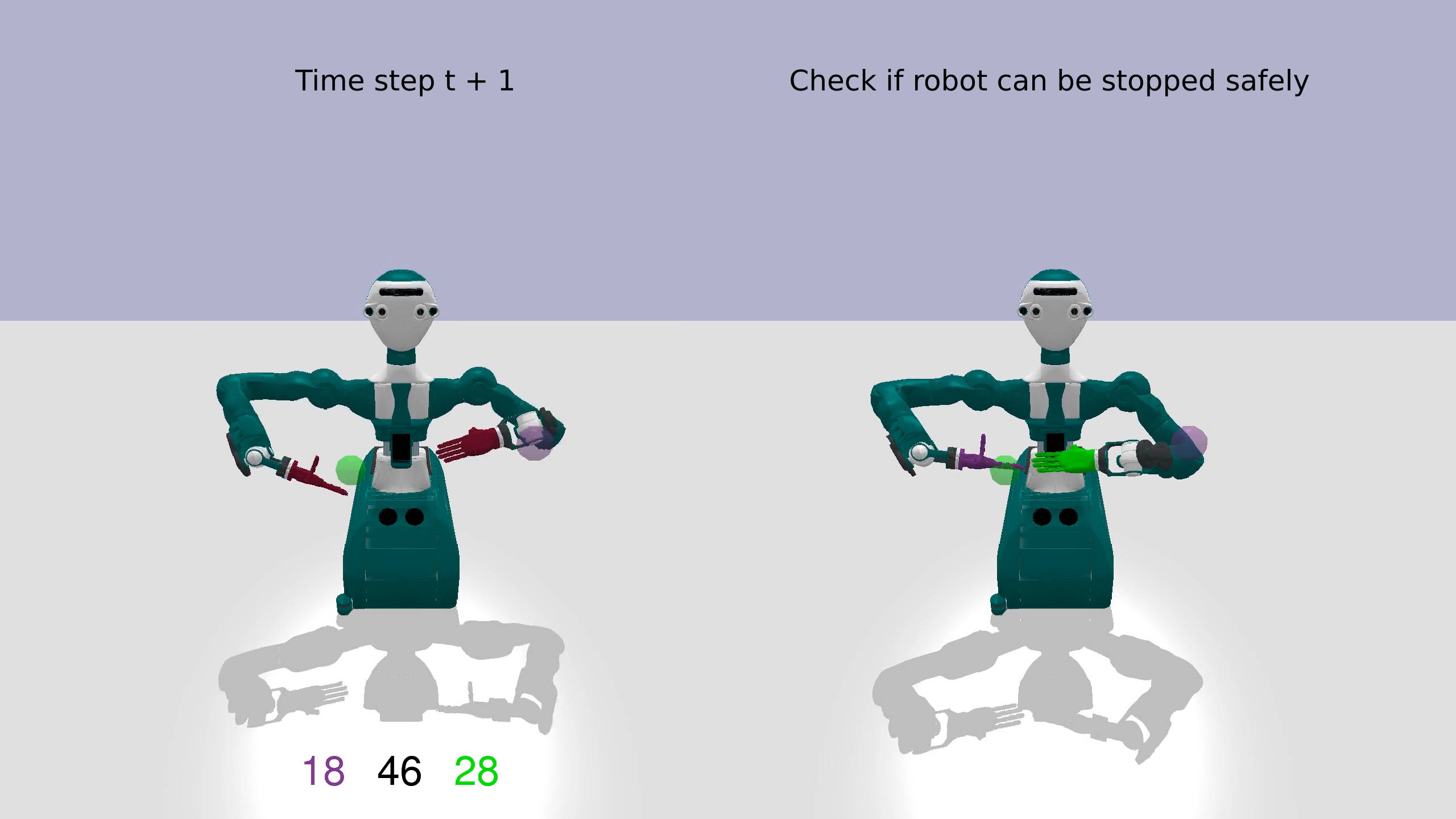}{$a_{t+3_B}''$}{0}{black}{}};
    \node[below of=t_0_brake_2, node distance=\distanceTopBottom, outer sep=1.5pt, yshift=0.0cm,  image_collection](t_1_brake_2){\timeStepImageSeries{1}{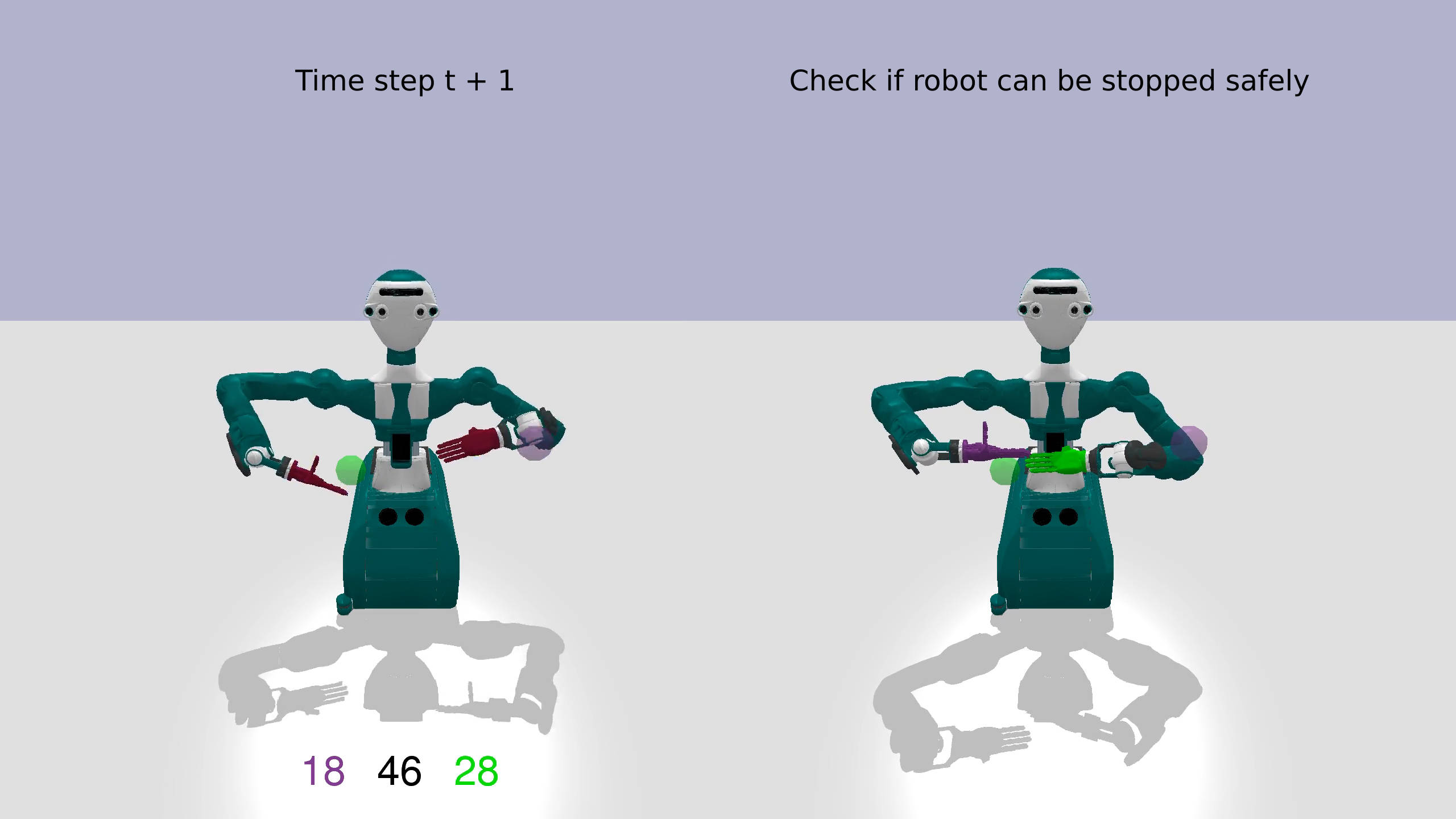}{$a_{t+4_B}''$}{1}{\colorCollision}{}};
    
    \node[below of=t_0_final, node distance=\distanceTopBottom, outer sep=1.5pt, yshift=0.0cm,  image_collection](t_1_final){\timeStepImageSeries{1}{figures/asb_t_0/right_brake_1_end_small.jpg}{$a_{t+2}$}{1}{\colorFinalSecond}{$a_{t+2_B}'$}};
    \path [] (t_1_brake_2.0) -- (t_1_final.180) node[pos=0.32, above=-0.0cm, scale=0.8, align=center]{\textcolor{\colorCollision}{Collision} \\ \textcolor{\colorCollision}{detected:}} node[pos=0.3, below=0.0cm, scale=0.85]{$a_{t+2} = a_{t+2_B}'$};
    
    \draw[] ($(t_0_brake_1.250) + (0cm, 0.2cm)$) -- + (0.0cm, -0.55cm) node[outer sep=0pt, inner sep=0pt](final_arrow_1){};
    \draw[] (final_arrow_1.center) -- + (5.0cm, 0.0cm) node[outer sep=0pt, inner sep=0pt](final_arrow_2){};
    \draw[arrow_style] (final_arrow_2.center) |- (t_1_final.180);
    
    \def\annotationHeight{0.1cm}
    \def\annotationScaling{0.88}
    \draw[color=\colorAnnotations] ($(t_1_action.225) + (0cm, -0.2cm)$) -- + (0.0cm, -1*\annotationHeight) node[outer sep=0pt, inner sep=0pt](background_simulation_left_corner){};
    \draw[color=\colorAnnotations] (background_simulation_left_corner.center) -- + (7.95cm, 0.0cm) node[pos=0.5, below=0.05cm, scale=\annotationScaling]{\textcolor{\colorAnnotations}{2. Background simulation of a backup trajectory}} node[outer sep=0pt, inner sep=0pt](background_simulation_right_corner){};
    \draw[color=\colorAnnotations] (background_simulation_right_corner.center) -- + (0.0cm, \annotationHeight);
    
    \draw[color=\colorAnnotations] ($(background_simulation_right_corner.center) + (0.95cm, 0.0cm)$) -- + (0.0cm, \annotationHeight) node[pos=0, outer sep=0pt, inner sep=0pt](trajectory_execution_left_corner){};
    \draw[color=\colorAnnotations] (trajectory_execution_left_corner.center) -- + (2.4cm, 0.0cm) node[pos=0.5, below=0.05cm, scale=\annotationScaling]{\textcolor{\colorAnnotations}{3. Motion execution}} node[outer sep=0pt, inner sep=0pt](trajectory_execution_right_corner){};
    \draw[color=\colorAnnotations] (trajectory_execution_right_corner.center) -- + (0.0cm, \annotationHeight);
    
    \draw[color=\colorAnnotations] ($(background_simulation_left_corner.center) + (-0.95cm, 0.0cm)$) -- + (0.0cm, \annotationHeight) node[pos=0, outer sep=0pt, inner sep=0pt](action_generation_right_corner){};
    \draw[color=\colorAnnotations] (action_generation_right_corner.center) -- + (-5.1cm, 0.0cm) node[pos=0.5, below=0.05cm, scale=\annotationScaling]{\textcolor{\colorAnnotations}{1. Action generation}} node[outer sep=0pt, inner sep=0pt](action_generation_left_corner){};
    \draw[color=\colorAnnotations] (action_generation_left_corner.center) -- + (0.0cm, \annotationHeight);
    
    \draw[color=\colorAnnotations] ($(t_0_action.135) + (-0.0cm, 0.3cm)$) -- + (0.0cm, 1*\annotationHeight) node[outer sep=0pt, inner sep=0pt](time_step_action_left_corner){};
    \draw[color=\colorAnnotations] (time_step_action_left_corner.center) -- + (2.20cm, 0.0cm) node[pos=0.5, scale=\annotationScaling, above=0.05cm]{\textcolor{\colorAnnotations}{Time interval $\Delta t_N$}} node[outer sep=0pt, inner sep=0pt](time_step_action_right_corner){};
    \draw[color=\colorAnnotations] (time_step_action_right_corner.center) -- + (0.0cm, -1*\annotationHeight);
    
    \draw[color=\colorAnnotations] ($(time_step_action_right_corner.center) + (0.15cm, 0.0cm)$) -- + (0.0cm, -1*\annotationHeight) node[pos=0, outer sep=0pt, inner sep=0pt](braking_trajectory_left_corner){};
    \path[name path=horizontalLine] (braking_trajectory_left_corner.center) -- + (9.0cm, 0.0cm); 
    \path[name path=verticalLineOne] (background_simulation_right_corner.center) -- + (0.0cm, 6.5cm); 
    \path [name intersections={of=horizontalLine and verticalLineOne, by=braking_trajectory_right_corner}];
    \draw[color=\colorAnnotations] (braking_trajectory_left_corner.center) -- (braking_trajectory_right_corner.center) node[pos=0.5, above=0.05cm, scale=\annotationScaling]{\textcolor{\colorAnnotations}{Braking trajectory: $n \cdot \Delta t_N$ with $n \in \mathbb{N}$}} node[outer sep=0pt, inner sep=0pt](braking_trajectory_right_corner){};
    \draw[color=\colorAnnotations] (braking_trajectory_right_corner.center) -- + (0.0cm, -1*\annotationHeight);
    
    \path[name path=verticalLineTwo] (trajectory_execution_left_corner.center) -- + (0.0cm, 6.5cm); 
    \path[name path=verticalLineThree] (trajectory_execution_right_corner.center) -- + (0.0cm, 6.5cm); 
    \path [name intersections={of=horizontalLine and verticalLineTwo, by=time_step_final_left_corner}];
    \path [name intersections={of=horizontalLine and verticalLineThree, by=time_step_final_right_corner}];
    \draw[color=\colorAnnotations] (time_step_final_left_corner.center) -- + (0.0cm, -1*\annotationHeight);
    \draw[color=\colorAnnotations] (time_step_final_left_corner.center) -- (time_step_final_right_corner.center) node[pos=0.5, scale=\annotationScaling, above=0.05cm]{\textcolor{\colorAnnotations}{Time interval $\Delta t_N$}};
    \draw[color=\colorAnnotations] (time_step_final_right_corner.center) -- + (0.0cm, -1*\annotationHeight);
    
\end{tikzpicture}

%% file: figures/action_space.tex
	\def\ymax{2.5}
	\def\ymin{-2.4}
	\def\xdelta{2.955}
	\def\xmax{3*\xdelta + 0.5}
	\def\amin{-2.2}
	\def\amax{2.2}
	
	\def\aonemaxA{1.3}
	\def\aonemaxB{1.1}
	\def\jmin{-2.6} 
	\def\aonemaxC{\amin-\jmin}
	\def\astarB{-2.0}
	\def\astarC{-0}
	\def\jmax{0.65}
	\def\njmaxFractionC{0.75}
	\def\vstarminusoneA{1.2}
	\def\vstarminusoneB{1.0}
	\def\vstarminusoneC{0.8}
	\def\vstarA{-1.2}
	\def\vstarB{-0.4}
	\def\vstarC{0.0}
	\def\tfive{(\x - 5*\xdelta) / \xdelta}
	
	\def\jmaxVis{\jmax*3}
	\def\astarBVis{\astarB*1}
	\def\astarCVis{\astarC*1}
	
	\def\cfiveB{(\vstarminusoneB - \vstarB) / (0.5 * \jmaxVis - \astarBVis)}
	
	\def\dfiveB{\vstarB * (1-\cfiveB)}
	
	\def\cfiveC{(\vstarminusoneC - \vstarC) / (0.5 * \jmaxVis - \astarCVis)}
	\def\dfiveC{\vstarC * (1-\cfiveC)}
	
	\def\steps{3}
	\def\showLimits{1}
	\def\violation{0}
	\def\showRedLine{1}
	\def\showAcc{1}
	
	\def\redLineLinestyle{solid}
    \def\limitsLinestyle{dashed}
	
	\def\azeroMax{1.6}
	\def\aoneMax{1.1}
	\def\atwoMax{0.9}
	\def\athreeMax{1.2}
	\def\afourMax{1.6}
	
	\def\azeroMin{-1.4}
	\def\aoneMin{-0.9}
	\def\atwoMin{-1.3}
	\def\athreeMin{-1.5}
	\def\afourMin{\amin}
	
	\def\azero{0.6}
	
	\def\actionColor{black!85}
	
	\def\aoneViolation{1.5}
	\ifnum \violation = 1
		\def\aone{\aoneViolation}
	\else
		\def\aone{0.5 * \aoneMin + 0.5 * \aoneMax}
	\fi
	
	\def\atwo{0.75 * \atwoMin + 0.25 * \atwoMax}
	\def\athree{1.0 * \athreeMax + 0.0 * \athreeMin}
	\def\afour{1.2}

	\definecolor{POS_LIM_A}{RGB}{0,120,0}%
	\definecolor{POS_LIM_B}{RGB}{0,0,150} %
	\definecolor{POS_LIM_C}{RGB}{0,120,0}%
	\definecolor{POS_LIM_D}{RGB}{214,39,40}
	
	\begin{tikzpicture}[scale=1, every node/.style={scale=1.1}]
	\draw [<-,thick] (0,\ymax) node (yaxis) [above] {$a$} -- (0,\ymin) node[below=0.1cm, name=nodet0] {$t$};
	
	\draw[dashed] (\xdelta, \ymax)  -- (\xdelta, \ymin) node[below=0.1cm] {$t+1$};
	\draw[dashed] (2*\xdelta, \ymax)  -- (2*\xdelta, \ymin) node[below=0.1cm] {$t+2$};
	\draw[dashed] (3*\xdelta, \ymax)  -- (3*\xdelta, \ymin) node[below=0.1cm] {$t+3$};
	
	\draw[dashed] (\xdelta, \ymax - 0.1)  -- + (\xdelta, 0) node[pos=0.5, fill=white, scale=0.9] {\textcolor{black!70}{$\Delta t_N$}};

	\ifnum \showLimits > 0
		\draw[\limitsLinestyle, color=POS_LIM_A, very thick] (0,\azeroMax) node[left=0.0975cm, color=black] {$a_{t_{max}}$} -- (\xdelta,\aoneMax); 
		\ifnum \steps > 1
			\draw[\limitsLinestyle, color=POS_LIM_A, very thick] (\xdelta,\aoneMax) -- (2*\xdelta,\atwoMax); 
			\ifnum \steps > 2
				\draw[\limitsLinestyle, color=POS_LIM_A, very thick] (2*\xdelta,\atwoMax) -- (3*\xdelta,\athreeMax); 
				\ifnum \steps > 3
					\draw[\limitsLinestyle, color=POS_LIM_A, very thick] (3*\xdelta,\athreeMax) -- (4*\xdelta,\afourMax);
					\fi
			\fi
		\fi
		\draw[\limitsLinestyle, color=POS_LIM_C, very thick] (0,\azeroMin) node[left=0.1475cm, color=black] {$a_{t_{min}}$} -- (\xdelta,\aoneMin); 
		\ifnum \steps > 1
			\draw[\limitsLinestyle, color=POS_LIM_C, very thick] (\xdelta,\aoneMin) -- (2*\xdelta,\atwoMin); 
			\ifnum \steps > 2
				\draw[\limitsLinestyle, color=POS_LIM_C, very thick] (2*\xdelta,\atwoMin) -- (3*\xdelta,\athreeMin); 
				\ifnum \steps > 3
					\draw[\limitsLinestyle, color=POS_LIM_C, very thick] (3*\xdelta,\athreeMin) -- (4*\xdelta,\afourMin);
				\fi
			\fi
		\fi
	\fi
	\ifnum \showRedLine = 1
	    \draw[\redLineLinestyle, color=POS_LIM_D, very thick] (0,\azeroMin)  -- (0,\azeroMax);
		\draw[\redLineLinestyle, color=POS_LIM_D, very thick] (\xdelta,\aoneMin)  -- (\xdelta,\aoneMax);
		\draw[\redLineLinestyle, color=POS_LIM_D, very thick] (2*\xdelta,\atwoMin)  -- (2*\xdelta,\atwoMax);
		\draw[\redLineLinestyle, color=POS_LIM_D, very thick] (3*\xdelta,\athreeMin)  -- (3*\xdelta,\athreeMax);
		\node[color=\actionColor, scale=0.9] at (0.5*\xdelta, \amin + 0.3) {$\boldsymbol{\underline{a}_t=0.0}$}; 
		\node[color=\actionColor, scale=0.9] at (1.5*\xdelta, \amin + 0.3) {$\underline{a}_{t+1}=-0.5$}; 
		\node[color=\actionColor, scale=0.9] at (2.5*\xdelta, \amin + 0.3) {$\underline{a}_{t+2}=1.0$}; 
	\fi

	\ifnum \showAcc = 1
		\ifnum \violation = 0
			\draw[color=POS_LIM_B, very thick] (0,\azero)  -- (\xdelta,\aone) node[pos=1.0, above=0.13cm, color=black, xshift=0.75cm, name=a_t_plus_one_n, outer sep=0pt, inner sep=1.25pt]{$\boldsymbol{a_{t+1_N}}$}; %
			\draw[-stealth] (a_t_plus_one_n.186) -- ($(\xdelta, \aone) + (0.03*\xdelta, 0.03*\xdelta)$);
			\fill[black] (\xdelta,\aone) circle (1.5pt);
    		\draw[|-stealth, thick, draw=black]  ($(\xdelta, \aoneMin) + (-0.065*\xdelta, 0)$) -- ($(\xdelta, \aone) + (-0.065*\xdelta, 0)$) node[pos=0.5, left=0.05cm, color=\actionColor, scale=0.9] {$\boldsymbol{\underline{a}_{t}}$};
		\else
			\draw[color=POS_LIM_B, very thick, dashed] (0,\azero)  -- (\xdelta,\aoneViolation) node[pos=0, left=0.15cm, color=black] {$a_0$}; %
		\fi
		\ifnum \steps > 1
			\ifnum \violation = 0
				\draw[color=POS_LIM_B, very thick] (\xdelta,\aone) -- (2*\xdelta,\atwo) node[pos=1.0, above=-0.53cm, color=black, xshift=0.65cm, name=a_t_plus_two, outer sep=0pt, inner sep=1.75pt]{};%
    			\fill[black] (2*\xdelta,\atwo) circle (1.5pt);
        		\draw[|-stealth, thick, draw=\actionColor]  ($(2*\xdelta, \atwoMin) + (-0.065*\xdelta, 0)$) -- ($(2*\xdelta, \atwo) + (-0.065*\xdelta, 0)$) node[pos=0.65, left=0.05cm, color=\actionColor, scale=0.9] {$\underline{a}_{t+1}$};
			\else
				\draw[color=POS_LIM_B, very thick, dashed] (\xdelta,\aoneViolation) -- (2*\xdelta,\atwo); 
			\fi
			
			\ifnum \steps > 2
				\draw[color=POS_LIM_B, very thick] (2*\xdelta,\atwo) -- (3*\xdelta,\athree); 
				\fill[black] (3*\xdelta,\athree) circle (1.5pt);
        		\draw[|-stealth, thick, draw=\actionColor]  ($(3*\xdelta, \athreeMin) + (-0.065*\xdelta, 0)$) -- ($(3*\xdelta, \athree) + (-0.065*\xdelta, -0.2)$) node[pos=0.5, left=0.05cm, color=\actionColor, scale=0.9] {$\underline{a}_{t+2}$};
				\ifnum \steps > 3
					\draw[color=POS_LIM_B, very thick] (3*\xdelta,\athree) -- (4*\xdelta,\afour);
				\fi
			\fi
		\fi
	\fi

	\fill[black] (0,\azero) circle (1.5pt) node[above=\azero, left=0.12cm, color=black] {$a_t$};

	\draw[dotted, thick] (0, \amin) node[left=0.23cm] {$a_{min}$}  -- (3*\xdelta, \amin) ;
	\draw[dotted, thick] (0, \amax) node[left=0.16cm] {$a_{max}$}  -- (3*\xdelta, \amax) ;

	\end{tikzpicture} 

%% file: root.bbl
\begin{thebibliography}{10}
\providecommand{\url}[1]{#1}
\csname url@rmstyle\endcsname
\providecommand{\newblock}{\relax}
\providecommand{\bibinfo}[2]{#2}
\providecommand\BIBentrySTDinterwordspacing{\spaceskip=0pt\relax}
\providecommand\BIBentryALTinterwordstretchfactor{4}
\providecommand\BIBentryALTinterwordspacing{\spaceskip=\fontdimen2\font plus
\BIBentryALTinterwordstretchfactor\fontdimen3\font minus
  \fontdimen4\font\relax}
\providecommand\BIBforeignlanguage[2]{{%
\expandafter\ifx\csname l@#1\endcsname\relax
\typeout{** WARNING: IEEEtran.bst: No hyphenation pattern has been}%
\typeout{** loaded for the language `#1'. Using the pattern for}%
\typeout{** the default language instead.}%
\else
\language=\csname l@#1\endcsname
\fi
#2}}

\bibitem{ShortPaperMotionSafety}
T.~{Fraichard}, ``A short paper about motion safety,'' in \emph{Proceedings
  2007 IEEE International Conference on Robotics and Automation}, 2007.

\bibitem{altman1999constrained}
E.~Altman, \emph{Constrained Markov decision processes}.\hskip 1em plus 0.5em
  minus 0.4em\relax CRC Press, 1999.

\bibitem{liu2021robot}
P.~Liu, D.~Tateo, H.~B. Ammar, and J.~Peters, ``Robot reinforcement learning on
  the constraint manifold,'' in \emph{5th Annual Conference on Robot Learning},
  2021.

\bibitem{achiam2017constrained}
J.~Achiam, D.~Held, A.~Tamar, and P.~Abbeel, ``Constrained policy
  optimization,'' in \emph{International Conference on Machine Learning}, 2017,
  pp. 22--31.

\bibitem{liu2020ipo}
Y.~Liu, J.~Ding, and X.~Liu, ``Ipo: Interior-point policy optimization under
  constraints,'' in \emph{Proceedings of the AAAI Conference on Artificial
  Intelligence}, vol.~34, no.~04, 2020, pp. 4940--4947.

\bibitem{cowen2020samba}
A.~I. Cowen-Rivers, D.~Palenicek, V.~Moens, M.~Abdullah, A.~Sootla, J.~Wang,
  and H.~Ammar, ``Samba: Safe model-based \& active reinforcement learning,''
  \emph{arXiv preprint arXiv:2006.09436}, 2020.

\bibitem{dalal2018safe}
G.~Dalal, K.~Dvijotham, M.~Vecerik, T.~Hester, C.~Paduraru, and Y.~Tassa,
  ``Safe exploration in continuous action spaces,'' \emph{arXiv preprint
  arXiv:1801.08757}, 2018.

\bibitem{alshiekh2018safe}
M.~Alshiekh, R.~Bloem, R.~Ehlers, B.~K{\"o}nighofer, S.~Niekum, and U.~Topcu,
  ``Safe reinforcement learning via shielding,'' in \emph{Thirty-Second AAAI
  Conference on Artificial Intelligence}, 2018.

\bibitem{pham2018optlayer}
T.-H. Pham, G.~De~Magistris, and R.~Tachibana, ``Optlayer-practical constrained
  optimization for deep reinforcement learning in the real world,'' in
  \emph{2018 IEEE International Conference on Robotics and Automation
  (ICRA)}.\hskip 1em plus 0.5em minus 0.4em\relax IEEE, 2018, pp. 6236--6243.

\bibitem{ames2014control}
A.~D. Ames, J.~W. Grizzle, and P.~Tabuada, ``Control barrier function based
  quadratic programs with application to adaptive cruise control,'' in
  \emph{53rd IEEE Conference on Decision and Control}.\hskip 1em plus 0.5em
  minus 0.4em\relax IEEE, 2014.

\bibitem{wang2017safe}
L.~Wang, A.~D. Ames, and M.~Egerstedt, ``Safe certificate-based maneuvers for
  teams of quadrotors using differential flatness,'' in \emph{2017 IEEE
  International Conference on Robotics and Automation (ICRA)}.

\bibitem{cheng2019end}
R.~Cheng, G.~Orosz, R.~M. Murray, and J.~W. Burdick, ``End-to-end safe
  reinforcement learning through barrier functions for safety-critical
  continuous control tasks,'' in \emph{Proceedings of the AAAI Conference on
  Artificial Intelligence}, vol.~33, no.~01, 2019, pp. 3387--3395.

\bibitem{anevlavis2021controlled}
T.~Anevlavis, Z.~Liu, N.~Ozay, and P.~Tabuada, ``Controlled invariant sets:
  implicit closed-form representations and applications,'' 2021.

\bibitem{pannocchi2021trust}
L.~{Pannocchi}, T.~{Anevlavis}, and P.~{Tabuada}, ``Trust your supervisor:
  quadrotor obstacle avoidance using controlled invariant sets,'' in \emph{2021
  IEEE/RSJ International Conference on Intelligent Robots and Systems
  (IROS)}.\hskip 1em plus 0.5em minus 0.4em\relax IEEE, 2021.

\bibitem{rubrecht2012motion}
S.~Rubrecht, V.~Padois, P.~Bidaud, M.~De~Broissia, and M.~D.~S. Simoes,
  ``Motion safety and constraints compatibility for multibody robots,''
  \emph{Autonomous Robots}, vol.~32, no.~3, pp. 333--349, 2012.

\bibitem{koller2018learning}
T.~Koller, F.~Berkenkamp, M.~Turchetta, and A.~Krause, ``Learning-based model
  predictive control for safe exploration,'' in \emph{2018 IEEE conference on
  decision and control (CDC)}.\hskip 1em plus 0.5em minus 0.4em\relax IEEE,
  2018.

\bibitem{tan2018sim}
J.~Tan, T.~Zhang, E.~Coumans, A.~Iscen, Y.~Bai, D.~Hafner, S.~Bohez, and
  V.~Vanhoucke, ``Sim-to-real: Learning agile locomotion for quadruped
  robots,'' \emph{arXiv preprint arXiv:1804.10332}, 2018.

\bibitem{gu2017deep}
S.~Gu, E.~Holly, T.~Lillicrap, and S.~Levine, ``Deep reinforcement learning for
  robotic manipulation with asynchronous off-policy updates,'' in \emph{2017
  IEEE international conference on robotics and automation (ICRA)}.\hskip 1em
  plus 0.5em minus 0.4em\relax IEEE, 2017, pp. 3389--3396.

\bibitem{kiemel2020learning}
J.~C. Kiemel and T.~Kröger, ``Learning robot trajectories subject to kinematic
  joint constraints,'' in \emph{2021 IEEE International Conference on Robotics
  and Automation (ICRA)}.\hskip 1em plus 0.5em minus 0.4em\relax IEEE, 2021.

\bibitem{faverjon1987local}
B.~Faverjon and P.~Tournassoud, ``A local based approach for path planning of
  manipulators with a high number of degrees of freedom,'' in
  \emph{Proceedings. 1987 IEEE international conference on robotics and
  automation}, vol.~4.\hskip 1em plus 0.5em minus 0.4em\relax IEEE, 1987, pp.
  1152--1159.

\bibitem{kiemel2020trueadapt}
J.~C. {Kiemel}, R.~{Weitemeyer}, P.~{Meißner}, and T.~{Kröger}, ``TrueÆdapt:
  Learning smooth online trajectory adaptation with bounded jerk, acceleration
  and velocity in joint space,'' in \emph{2020 IEEE/RSJ International
  Conference on Intelligent Robots and Systems (IROS)}, 2020.

\bibitem{coumans2016pybullet}
E.~Coumans and Y.~Bai, ``Pybullet, a python module for physics simulation for
  games, robotics and machine learning,'' 2016.

\bibitem{asfour2018armar}
T.~Asfour, L.~Kaul, M.~W{\"a}chter, S.~Ottenhaus, P.~Weiner, S.~Rader,
  R.~Grimm, Y.~Zhou, M.~Grotz, F.~Paus, \emph{et~al.}, ``Armar-6: A
  collaborative humanoid robot for industrial environments,'' in \emph{2018
  IEEE-RAS 18th International Conference on Humanoid Robots (Humanoids)}.

\bibitem{schulman2017proximal}
J.~Schulman, F.~Wolski, P.~Dhariwal, A.~Radford, and O.~Klimov, ``Proximal
  policy optimization algorithms,'' \emph{arXiv preprint arXiv:1707.06347},
  2017.

\bibitem{kiemel2022path}
J.~C. {Kiemel} and T.~{Kröger}, ``Learning time-optimized path tracking in
  joint space with or without sensory feedback,'' in \emph{2022 IEEE/RSJ
  International Conference on Intelligent Robots and Systems (IROS)}, 2022.

\end{thebibliography}
